%% file: paper-main.tex
\PassOptionsToPackage{table}{xcolor}

\documentclass[preprint,12pt]{elsarticle}
\usepackage[table,xcdraw]{xcolor}
\usepackage[numbers]{natbib}
\usepackage{multirow}
\usepackage{algorithmic}
\usepackage{graphicx}
\usepackage{textcomp}
\usepackage{mathtools}
\usepackage{url}
\usepackage{amsmath}
\usepackage{tabularx}
\usepackage{amssymb}
\usepackage{todonotes}
\usepackage[ruled,linesnumbered]{algorithm2e}
\usepackage{float}
\restylefloat{table}
\usepackage{caption}
\usepackage{subcaption}
\makeatletter
\def\ps@pprintTitle{%
  \let\@oddhead\@empty
  \let\@evenhead\@empty
  \let\@oddfoot\@empty
  \let\@evenfoot\@oddfoot
}
\makeatother

\usepackage{amssymb}

\journal{}

\begin{document}

\begin{frontmatter}



\title{A new algorithm for Subgroup Set Discovery based on Information Gain}


\author[inst1]{Daniel Gómez-Bravo}

\affiliation[inst1]{organization={Centro de Tecnología Biomédica},
            addressline={Crta. M40, Km. 38}, 
            city={Pozuelo de Alarcón},
            postcode={28223}, 
            state={Madrid},
            country={Spain}}

\affiliation[inst3]{organization={Escuela Técnica Superior de Ingenieros Informáticos},
            addressline={C. de los Ciruelos}, 
            city={Boadilla del Monte},
            postcode={28660}, 
            state={Madrid},
            country={Spain}}

\author[inst1]{Aaron García}
\author[inst1,inst3]{Guillermo Vigueras}
\author[inst1,inst3]{Belén Ríos}
\author[inst2]{Mariano Provencio}
\author[inst1,inst3]{Alejandro Rodríguez-González}

\affiliation[inst2]{organization={Hospital Universitario Puerta de Hierro Majadahonda},
            addressline={C. Joaquín Rodrigo, 1}, 
            city={Majadahonda},
            postcode={28222}, 
            state={Madrid},
            country={Spain}}

\begin{abstract}

Pattern discovery is a machine learning technique that aims to find sets of items, subsequences, or substructures that are present in a dataset with a higher frequency value than a manually set threshold. When dealing with sequential data, a frequent subsequence represents a pattern that occurs regularly in the sequence of items. On the other hand, a substructure can take various structural forms, such as subgraphs, subtrees, or sublattices, which can be combined with itemsets or subsequences. If a substructure occurs frequently in a database, it is known as a (frequent) structural pattern. This process helps to identify recurring patterns or relationships within the data, allowing for valuable insights and knowledge extraction.

In this work, we propose Information Gained Subgroup Discovery (IGSD), a new SD algorithm for pattern discovery that combines Information Gain and Odds Ratio (OR) as a multi-criteria for pattern selection. The algorithm tries to tackle some limitations of state-of-the-art SD algorithms like the need for fine-tuning of key parameters for each dataset, usage of a single pattern search criteria set by hand, usage of non-overlapping data structures for subgroup space exploration, and the impossibility to search for patterns by fixing some relevant dataset variables. Thus, we compare the performance of IGSD with two state-of-the-art SD algorithms: FSSD and SSD++. 

Performance comparison of FSSD, SSD++ and IGSD is done by finding patterns using these three algorithms in eleven datasets. For the performance evaluation, we also propose to complement standard SD measures with some metrics like Information Gain, Odds Ratio, and p-value, not considered typically in SD literature. Obtained results show that FSSD and SSD++ algorithms provide less reliable patterns, due to a lower confidence value, and also provide reduced sets of patterns when compared with the proposed IGSD algorithm for all datasets considered. Additionally, IGSD provides better OR values than FSSD and SSD++, stating a higher dependence between patterns and targets.

Moreover, patterns obtained for one of the datasets used, have been validated by a group of domain experts. Thus, patterns provided by IGSD show better agreement with experts than patterns obtained by FSSD and SSD++ algorithms. The results presented demonstrate the suitability of the proposed IGSD algorithm as a method for pattern discovery and suggest that the inclusion of non-standard SD metrics allows to better evaluate discovered patterns.

\end{abstract}


\end{frontmatter}


\input{tex/intro.tex}

\input{tex/methodology.tex}

\input{tex/clinical_guidelines.tex}

\input{tex/results.tex}

\input{tex/conclusions.tex}

\appendix


\bibliographystyle{elsarticle-num} 
\bibliography{AIIMBib}





\end{document}

%% file: tex/intro.tex
\section{Introduction}
\label{sec:introduction}

Pattern discovery or pattern mining is a machine learning technique that aims to find sets of items, subsequences, or substructures that are present in a dataset with a higher frequency value than a manually set threshold. In this context, a set of items that frequently appear together in a transaction data set, (e.g. milk and bread), is referred to as a frequent itemset. When dealing with sequential data, a frequent subsequence represents a pattern that occurs regularly in the sequence of items. On the other hand, a substructure can take various structural forms, such as subgraphs, subtrees, or sublattices, which can be combined with itemsets or subsequences. If a substructure occurs frequently in a graph database, it is known as a (frequent) structural pattern. This process helps identify recurring patterns or relationships within the data, allowing for valuable insights and knowledge extraction \cite{han_frequent_2007}. Also known as frequent pattern mining, it was initially popularized for market basket analysis, especially in the form of association rule mining. In this way, customer buying habits can be examined by identifying associations between different items that customers place in their shopping baskets. For instance, the analysis may reveal that customers who buy milk are highly likely to purchase cereal during the same shopping trip, and it can further specify which types of cereal are commonly associated with milk purchases \cite{agrawal_mining_nodate}. 

Among machine learning methods for pattern search, Subgroup Discovery (SD) \cite{herrera_overview_2011}  has been used previously to find relevant patterns in datasets. SD is a data mining task that aims to identify and extract interpretable patterns from the data, which exhibit interesting or exceptional characteristics with respect to a specific property of interest. It has been used in several fields, such as clinical applications \cite{esnault_q-finder_2020, subgroup_2023}, and technical applications \cite{atzmueller_subgroup_2015}, among others. However, several limitations were found due to the low complexity of the rules. Therefore, in this work, we propose the use of a wider range of features from datasets to discover more personalized patterns. 


Even if SD has been proposed as a useful technique, the standard version and state of art implementations \cite{belfodil_fssd_2019,proenca_discovering_2021} present several limitations. First, fine-tuning some key parameters is always necessary (i.e. \textit{Beam width}) for each analyzed dataset. Moreover, regarding the discovery of patterns, they are usually obtained by maximizing some index such as weighted relative accuracy (\textit{WRAcc}). However, pattern complexity is not optimized, and obtained patterns may lack interesting information. Additionally, previous algorithms do not offer the option of fixing some important dataset variables to be present in the discovered patterns, thus reducing the interest or acceptance of patterns by problem domain experts. Furthermore, subgroup lists are used in most recent SD algorithms and this can pose a problem as some information can be lost in non-overlapping subgroups found in datasets. Finally, the quality of discovered patterns is evaluated based on single index criteria, and sometimes the selection of evaluation indices is not unified across different SD algorithms.

Domain expert validation is a key point to find interesting conclusions on SD analysis. In the CN2-SD study, \cite{lavrac2004subgroup}, validation was performed for SD results on a real-life problem of traffic accident analysis dataset. Thus, interpretable and relevant patterns obtained from SD algorithms are necessary. However, SD literature typically lack validation of results by a set of experts. 


To overcome these issues, in this work, we propose a new SD algorithm, InfoGained-SD (IGSD). This algorithm searches patterns through an optimization that combines information gained \cite{noda_discovering_1999,prasetiyowati_determining_2021} and odds ratio \cite{szumilas_explaining_2010,dominguez-lara_odds_2018} metrics. In addition, no fine-tuning regarding \textit{Beam width} is required by the user. Furthermore, it allows fixing key attributes in discovered patterns by an expert in the field of study, in order to increase acceptance of discovered patterns. Besides, a subgroup set is obtained, so no information is lost for the analyzed input data. Thus, this algorithm is used to examine different datasets in order to find relevant patterns, taking into account characteristics that might be relevant for expert validation. Multiple datasets with heterogeneous variable types (categorical, numerical, and mixed) are assessed in this study, to prove that our method can be useful in the pattern discovery data mining field, obtaining relevant and interpretable patterns, for a domain expert.

\section{State of the Art}

Subgroup Discovery (SD) \cite{herrera_overview_2011} has been proven as a suitable method for identifying statistical and  relevant patterns in datasets. It has been applied to clinical trials, precision medicine, and treatment optimization or disease study \cite{zhang_subgroup_2018,loh_subgroup_2019, korepanova_subgroup_2018,subgroup_2023,ibald-mulli_identification_nodate}. Other  applications are found in the bibliography, such as in social media study and \cite{atzmueller_mining_2012} and smart electricity meter data \cite{jin_subgroup_2014}, among other fields.

Even if SD has been proven as a suitable method for pattern discovery, an ongoing problem with this data mining technique is the difficulty of interpreting or analyzing the results produced, either because of the complexity and the large amount of information or because of their relevancy \cite{sun_is_2010,burke_three_2015}. In order to reduce the number of results obtained, some solutions such as ranking and selecting the best $n$ associations, eliminating associations composed of many features, or discarding associations with a specific measure value below a manual threshold have been proposed in the literature. Thus, SSD++ \cite{proenca_robust_2022} and FSSD \cite{belfodil_fssd_2019} seem to lead the state of the art of SD algorithms. SSD++ relies on beam search strategy, a heuristic approach for discovering subgroups in a population. This process, during the exploration phase, looks through combinations of variables until the maximum search depth of the dataset is covered, and stores only a predetermined number of subgroups at each level (\textit{Beam width}), which are the ones having the best heuristic cost. This means that the same value of \textit{Beam width} can result in a too large or insufficient number of patterns, depending on the field of study, avoiding to obtain relevant patterns.


On the other hand, the exploration of subgroups in FSSD is done using
the DEPTH-FIRST-SEARCH strategy, which does not have the limitation of discarding patterns when exploring all possible combinations, however, it is still necessary to determine the optimal number of patterns to return, and this poses a problem as described before. Therefore, using a manual threshold does not seem to be a suitable technique to find the most relevant patterns, and removing the necessary fine-tuning of key parameters such as \textit{Beam width} seems to be a well-suited approach in this field. Additionally, even if a correct threshold is manually set, found patterns should represent a balance between the complexity of patterns and dependency on the target variable. Actual approaches of the state of the art are not considering this concept.

Regarding exploration strategy of subgroups search space, both SSD++ and FSSD are based on subgroup lists, which can be defined as the fragmentation of the subgroup population into multiple sections, each of which is represented by a unique group. This feature is not in line with the aim of IGSD, which is to discover how different characteristics may influence each instance in a dataset. Hence, some information may be lost if subgroup lists are used. Also, validation of obtained patterns is an important aspect in SD, since an expert can evaluate the significance of obtained patterns or the relevant variables that should be present in patterns. Thus, SSD++ and FSSD do not allow to specify a set of key attributes to be present in returned patterns. 

Finally, several performance indices have been used in the literature to compare different SD algorithm executions (such as for SSD++ and FSSD). However, there is a lack of homogeneity in the indices used and a complete comparison regarding all the measures should be provided to better analyze the performance of the different approaches.



%% file: tex/methodology.tex
\RestyleAlgo{ruled}
\section{Materials and Methods}

\subsection{Data description}
\label{sec:data-desc}

The objective of this study is to evaluate and contrast the proposed IGSD algorithm with existing state-of-the-art algorithms like FSSD and SSD++. To accomplish this, a total of 10 datasets were chosen from the UCI and Mulan repositories, along with the P4Lucat dataset.
In Table \ref{tab:datasets}, the datasets were classified into three categories based on their data type: numeric datasets are represented in green, nominal datasets in blue, and mixed datasets containing both numeric and nominal data in yellow. Furthermore, the "Rows" column shows the number of records in each dataset, the "Targets" column indicates the number of targets for each dataset along with the number of possible values, and the "DataType" column specifies the count of nominal and/or numeric columns present in each dataset.
Furthermore, P4Lucat and Genbase datasets present more than one target, so it was decided to transform all the possible targets into one unique target, using the combination of the different target values for each dataset. Thus, resulting in one target option for the P4Lucat dataset with 4 possible values and one target option for the Genbase dataset with 32 possible values. This transformation strategy (One-vs-Rest) for handling multi-target datasets, and obtain a single target is explained later in Section~\ref{sec:SD}.

\begin{table}[H]
\centering
\begin{tabular}{|c|c|c|c|c|}
\hline
\textbf{Type}  & \textbf{Dataset}                               & \textbf{Rows} & \textbf{Targets} & \textbf{DataType(nom/num)} \\ \hline \hline
\cellcolor[HTML]{92D050} & \cellcolor[HTML]{92D050}\textbf{Iris}          & 150           & 1 (3)            & 0/4                 \\ \cline{2-5} 
\cellcolor[HTML]{92D050} & \cellcolor[HTML]{92D050}\textbf{Echo}          & 108           & 1 (2)            & 0/6                 \\ \cline{2-5} 
\cellcolor[HTML]{92D050} & \cellcolor[HTML]{92D050}\textbf{Heart}         & 270           & 1 (2)            & 0/13                \\ \cline{2-5} 
\multirow{-4}{*}{\cellcolor[HTML]{92D050}\rotatebox{90}{Numeric}} & \cellcolor[HTML]{92D050}\textbf{Magic}         & 19020         & 1 (2)            & 0/10                \\ \hline \hline
\cellcolor[HTML]{8EA9DB} & \cellcolor[HTML]{8EA9DB}\textbf{tic-tac-toe}   & 958           & 1 (2)            & 9/0                 \\ \cline{2-5}
\cellcolor[HTML]{8EA9DB} & \cellcolor[HTML]{8EA9DB}\textbf{Vote}          & 435           & 1 (2)            & 16/0                \\ \cline{2-5}
\cellcolor[HTML]{8EA9DB} & \cellcolor[HTML]{8EA9DB}\textbf{P4Lucat}       & 650           & 2 $\to$ 1(4)            & 9/0                 \\ \cline{2-5}
\multirow{-4}{*}{\cellcolor[HTML]{8EA9DB}\rotatebox{90}{Mixed}} & \cellcolor[HTML]{8EA9DB}\textbf{Genbase}       & 662           & 27 $\to$ 1(32)          & 1186/0              \\ \hline \hline
\cellcolor[HTML]{FFD966} & \cellcolor[HTML]{FFD966}\textbf{Adult}         & 45222         & 1 (2)            & 8/6                 \\ \cline{2-5}
\cellcolor[HTML]{FFD966} & \cellcolor[HTML]{FFD966}\textbf{Nursery}       & 12960         & 1 (5)            & 7/1                 \\ \cline{2-5}
\multirow{-3}{*}{\cellcolor[HTML]{FFD966}\rotatebox{90}{Nominal}} & \cellcolor[HTML]{FFD966}\textbf{Breast-cancer} & 286           & 1 (2)            & 8/1                 \\ \hline
\end{tabular}
\caption{Datasets description}
\label{tab:datasets}
\end{table}

\subsection{Pattern discovery methods}
\label{sec:ml-methods}

In this section, we will discuss the methodologies employed for identifying patterns in the aforementioned data. Initially, we will provide an overview of SD and present essential definitions related to this field. Subsequently, we will introduce the IGSD algorithm, which is the proposed method for pattern discovery.

\subsubsection{Subgroup Discovery}
\label{sec:SD}

Subgroup Discovery (SD) is a data mining technique that aims to uncover meaningful associations between variables in relation to a specific property of interest \cite{sammut_encyclopedia_2017}. The literature distinguishes two versions or cultures of SD: Subgroup Identification (SI) and Knowledge Discovery in Databases (KDD) \cite{esnault_q-finder_2020}. In this study, the KDD culture is adopted due to its domain-agnostic nature, which allows for the utilization of diverse quality metrics or measures such as coverage, support, unusualness, and more. By employing these metrics, KDD endeavors to identify statistically significant subgroups that satisfy a given target property. \\

The following set of definitions is presented as a foundational background for key concepts that are common to SD algorithms: \\

\textbf{Dataset}: A dataset (D) can be defined as the set of items $I=(X,Y)$, where $X=\{k1-v1,k2-v2,..,kn-vn\}$ represents the conjunction of $attributes(k)-values(v)$ pairs and $Y$ the target value selected. The attributes set $(k)$ encompasses all the explanatory variables present in the dataset. The values $(v)$ can be classified into three types: numeric, boolean, and nominal. On the other hand, in literature, it is found that SD can be employed for binary, nominal and numerical targets, as stated in \cite{proenca_robust_2022}. SSD++ is capable of handling all types of targets while FSSD is limited to binary targets. On the other hand, regarding IGSD algorithm, the decision employed to handle the numeric targets is to transform them into nominal targets. Moreover, nominal targets will be treated as binary targets employing the One-vs-Rest strategy explained later on. \\

\textbf{Subgroup}: A subgroup (s) refers to a combination (Comb) of properties or features, which are attribute-value pairs that describe a distribution with respect to the Target$_{value}$ in a given dataset. Therefore, the properties or features (Comb) must contain a combination that exists in the dataset. Additionally, each attribute-value pair, also known as a selector, consists of an attribute, a condition, and a value. The possible conditions depend on the variable type: numeric variables support greater and less than \{$\geq$, $\leq$\} while binary and categorical support equal to \{==\}, i.e $\{attr1=="possible \; value"\}$ or $\{attr1\geq5\}$. These subgroups can be represented as individual patterns being regularly defined as:
\begin{equation}
s : Comb \rightarrow Target_{value}
\label{eq:Rule format}
\end{equation}

\textbf{Sets and Lists of Subgroups}: Subgroup sets can be described as disjunctions of subgroups, allowing for overlapping between subgroups within the same set. Against that, subgroup lists do not permit overlapping, meaning that each attribute of a subgroup is not contained within another subgroup. This distinction is crucial when comparing the performance of FSSD, SSD++, and IGSD algorithms.\\ 

\textbf{Quality Function}: A quality function \(q: \Omega_{sd} \times \Omega_E \rightarrow \mathbb{R}\) is employed to assess the effectiveness of a subgroup description \(sd\) belonging to the set \(\Omega_{sd}\), given a target concept \(t \in \Omega_E\), and to rank the discovered subgroups during the search process. Quality functions are presented here in a general context for subgroup discovery and will be subsequently elaborated upon as descriptive and predictive measures.

For binary target variables, various significant quality functions can be defined in the following form:

\[
q_a = n^a \cdot (p - p_0), \quad a \in [0, 1]
\]

Here, \(p\) represents the relative frequency of the target variable within the subgroup, \(p_0\) denotes the relative frequency of the target variable in the total population, and \(n\) indicates the size of the subgroup. The parameter \(a\) allows for a trade-off between the increase in the target share \(p - p_0\) and the generality \(n\) of the subgroup.\\

Regarding the target space \(\Omega_E\) a combination of scenarios can be found with a binary class target variable, a multi-class target variable (containing more than 2 possible values) resulting in a single-target analysis; and a combination of different target variables previously defined resulting in a multi-target problem. 

When it comes to multi-class dataset analysis, the literature offers various approaches for handling multi-class problems. For instance, SSD++ executes the SD algorithm for each target and incorporates only subgroups that enhance the information in a subgroup list. In the case of a multi-class problem with more than two classes, it can be transformed into a binary problem by employing the One-vs-Rest (OvR) strategy \cite{buntine_evaluation_2009}. Each class under consideration is treated as one class, while the remaining classes are grouped together as another class (not belonging to the class under study). This transformation allows binary target datasets to be analyzed effectively, and it is the procedure employed by IGSD to manage multi-class datasets.

Moreover, more than one variable of interest may be used in the analysis, thus a multi-target problem appears. These scenarios are explored in SSD++, where a subgroup list model is generated, considering the categorical distribution for each class found within those targets. The solution proposed in this work to manage the multi-target scenario is to generate a new target variable where all target variables and their respective classes are combined so that they are linked through a conjunction operation. This procedure must be done as a preprocessing step of the data before employing the IGSD algorithm. For example, in P4Lucat dataset we have two binary variables as targets: disease progression-relapse and toxicity. Consequently, the target variable will contain the information from both binary variables, namely \(Progression-Relapse\)=[YES/NO] and \(Toxicity\)=[YES/NO]. This combination results in a new target variable with four distinct classes.

The following paragraphs describe the main descriptive measures commonly found in the literature on SD. These measures allow for the evaluation of individual subgroups, enabling the comparison of results across different algorithms:

\begin{itemize}
\item Coverage \cite{herrera_overview_2011}: It measures the percentage of examples covered on average. This can be computed as:
\begin{equation}
Cov(R) = \frac{n(Cond)}{ns}
\label{eq:Cov}
\end{equation}
where \(ns\) is the number of total examples and \(n(Cond)\) is the number of examples that satisfy the conditions determined by the antecedent part of the pattern. The average coverage of a subgroup set is computed as:
\begin{equation}
COV = \frac{1}{nR} \sum_{i=1}^{nR} Cov(R_i)
\label{eq:COV}
\end{equation}
where \(nR\) is the number of induced patterns.

\item Confidence \cite{herrera_overview_2011}: It measures the relative frequency of examples satisfying the complete pattern among those satisfying only the antecedent. This can be computed as:
\begin{equation}
Cnf(R) = \frac{n(Target_{value}Cond)}{n(Cond)}
\label{eq:Conf}
\end{equation}
where \(n(Target_{value}Cond) = TP\) and it is the number of examples that satisfy the conditions and also belong to the value for the target variable in the pattern. The average confidence of a pattern set is computed as:
\begin{equation}
CNF = \frac{1}{nR} \sum_{i=1}^{nR} Cnf(R_i)
\label{eq:CONF}
\end{equation}

\item Size: The pattern set size is computed as the number of patterns in the induced pattern set.

\item Complexity: It measures the level of information presented in patterns. It is determined as the number of variables contained in the pattern.

\item Unusualness \cite{herrera_overview_2011}: This measure is described as the weighted relative accuracy of a pattern. It can be calculated as:
\begin{equation}
WRAcc(R) = Cov(R) * (Cnf(R) - \frac{n(Target_{value})}{ns} )
\label{eq:Wraacc}
\end{equation} 
The unusualness of a pattern can be described as the balance between its coverage, represented by \(Cov(R)\), and its accuracy gain, denoted by \(Cnf(R) - \frac{n(Target_{value})}{ns}\).
The average unusualness of a pattern set can be computed as:
\begin{equation}
WRAcc = \frac{1}{nR} \sum_{i=1}^{nR} WRAcc(Ri),
\label{eq:WRACC}
\end{equation}
\end{itemize}

In addition to the descriptive metrics discussed earlier, predictive measures can also be utilized to evaluate a pattern set, treating a set of subgroup descriptions as a predictive model. Although the primary objective of pattern discovery algorithms is not accuracy optimization, these measures can be employed to compare predictive performance.
\begin{itemize}
\item Predictive accuracy \cite{jin_huang_using_2005}: Predictive accuracy refers to the percentage of correctly predicted instances. In the case of a binary classification problem, the accuracy of a pattern set can be computed as:
\begin{equation}
ACC = \frac{TP+TN}{TP+TN+FP+FN}
\label{eq:ACC}
\end{equation}
where TP represents true positives, TN denotes true negatives, FP represents false positives, and FN denotes false negatives.
\end{itemize}

In this paper, we also incorporate quality functions that describe relevant aspects of patterns. One such measure is Information Gain (IG) \cite{noda_discovering_1999} \cite{prasetiyowati_determining_2021}, which quantifies the reduction in entropy or surprise by splitting a dataset based on a specific value of a random variable. It is calculated as follows:

\[
IG(D, v) = H(D) - H(D | v)
\]

Here, \(IG(D, v)\) represents the information gain for the dataset \(D\) with respect to the variable \(v\), \(H(D)\) is the entropy of the dataset before any change, and \(H(D | v)\) is the conditional entropy of the dataset when the variable \(v\) is added.

The entropy of a dataset can be understood in terms of the probability distribution of observations within the dataset belonging to different classes. Thus, the entropy measures the level of uncertainty or randomness in the distribution of classes within the dataset. For example, in a binary classification problem with two classes, the entropy of a data sample can be calculated using the following formula:
\begin{equation}
Entropy = -(p(a) * \log(P(a)) + p(1-a) * \log(P(1-a)))
\label{eq:Entropy}
\end{equation}

In addition, we also employed the odds ratio (OR) measure\cite{szumilas_explaining_2010,dominguez-lara_odds_2018}, which is represents the association between an antecedent and an outcome. The OR represents the ratio of the odds of the outcome occurring given a specific antecedent, compared to the odds of the outcome occurring in the absence of that antecedent.

In this work, we utilize ORs to compare the relative odds of the occurrence of the outcome of interest based on specific patterns that contain multiple selectors. This measure enables us to evaluate the strength of the association between the antecedent (pattern) and the outcome of interest.
Consequently, odds ratios (ORs) can be utilized to assess whether adding a new selector to a pattern serves as a risk factor for a specific outcome. They also allow for comparing the magnitude of various risk factors associated with that outcome. This comparison helps determine the effectiveness of adding more information to a pattern.

In IGSD, ORs are employed as an index to select the most relevant subgroups based on the association between the antecedent and the target. By considering the ORs, IGSD identifies subgroups with higher odds ratios, indicating stronger associations between the antecedent and the target outcome. This selection process helps prioritize the most relevant subgroups in terms of their predictive power and relevance to the target.
The odds ratio (OR) can be calculated using the following formula:

\[
OR = \frac{TP \cdot TN}{FP \cdot FN}
\]

To interpret the OR as a size effect in \cite{dominguez-lara_odds_2018} is proposed the transformation of OR into Cohen's \(d\). This transformation makes the interpretation of the OR easier, as it allows for considering an $OR > 6.71$ to have a similar effect size, regardless of the actual magnitude of the OR.

In cases where Cohen's \(d\) is not obtained, comparing subgroup sets based solely on mean values of the OR may lead to distorted results. Higher OR values can disproportionately influence the mean, while lower OR values may not receive due consideration. To address this, four intervals are defined:

\begin{itemize}

\item \(OR < 1.68\) represents a very low effect.
\item  \(1.68 < OR < 3.47\) represents a low effect.
\item  \(3.47 < OR < 6.71\) represents a moderate effect.
\item  \(OR > 6.71\) represents a high effect.

\end{itemize}

For ease of numerical representation, a value is assigned to each interval, resulting in the odds ratio range (ORR) being defined from 1 to 4. This allows for a more balanced comparison between subgroups and avoids overemphasizing the impact of extremely high OR values.

Finally, we have employed the p-value as a subgroup filtering criterion, which is calculated using the Chi-Square statistical test \cite{mchugh_chi-square_2013}. A p-value threshold of 0.05 is commonly used as the standard criterion for statistical significance.

\subsection{IGSD algorithm}
\label{sec:IGSD} 

This section presents IGSD, a pattern discovery algorithm that aims to minimize pattern complexity while simultaneously maximizing the quality of the knowledge derived from discovered patterns.
This algorithm combines IG and ORR to identify, on the basis of IG, the attributes with greater relevance and, on the basis of ORR, the set of variable values that have a stronger dependence on a particular target.


As previously stated, the proposed algorithm attempts to overcome some limitations of current SD methods. As a result, prior algorithms necessitated adjusting key parameters for each dataset being analyzed. As a result, parameters like beam width, which affect discovered patterns and control the size of the search space, must be defined for each input dataset. In addition, previous algorithms tried to find patterns by maximizing a single index, usually weighted relative accuracy (\textit{WRAcc}), which necessitated manually setting a threshold for the optimization index for each analyzed dataset once more. Additionally, previous algorithms explored subgroup search space by making use of non-overlapping data structures like subgroup lists. Since non-overlapping information in explored subgroups can prevent the discovery of relevant and intriguing patterns, this can be a limitation. Additionally, some crucial dataset variables cannot be fixed to be present in the discovered patterns using previous algorithms. However, because experts in the field may require them to consider a pattern to be useful or interesting, patterns with fixed key variables are an important aspect.
Lastly, the quality of discovered patterns is evaluated using a single index, and the evaluation indices chosen by various SD algorithms may not always be consistent.

The IGSD algorithm addresses all of these limitations. As a result, this new strategy employs a dynamic threshold using IG as a single optimization index when searching for subgroups. This threshold will be used to select which selectors will be considered relevant options in each subgroup discovery step. Furthermore, there is no need to manually define an arbitrary value for this threshold. Instead, at each algorithm discovery step, it is dynamically calculated and modified for each explored subgroup. Since the IG threshold is dynamically adjusting the size of the search space, it is unnecessary to fine-tune the \textit{Beam width} parameter at this time. In addition, IGSD provides a uniform measure output that can be compared to that of other implementations.

Three arguments are needed to start the IGSD algorithm. These arguments can be used to choose between different options for the algorithm and do not require any fine-tuning. The arguments are the maximum depth during the exploration phase ($d_{max}$), the condition attributes ($Cond_{list}$), and the threshold mode ($t_{mode}$). The algorithm will use either the maximum IG threshold or the dynamic IG threshold, which is the default, according to the $t_{mode}$ variable.


The $d_{max}$ parameter determines the depth of the exploration space, which can also be interpreted as the pattern complexity or the maximum number of selectors that the patterns will have. In addition, the user can specify some dataset variables that must be present in the obtained patterns using the $Cond_{list}$ parameter.

\begin{figure}[H]
\centering
\includegraphics[width=\columnwidth, height=2.2in]{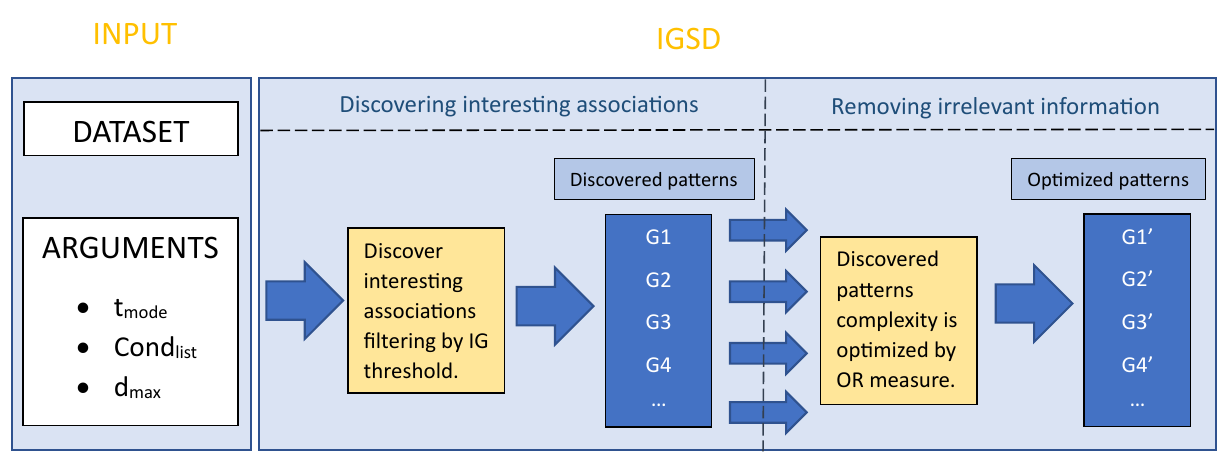}
\caption{Algorithm workflow}
\label{workflow}
\end{figure}


The algorithm's workflow is depicted in Figure \ref{workflow}, which shows the steps the algorithm takes. Finding interesting associations and removing irrelevant information from associations are two tasks that can be defined. First, a dataset and the values of the parameters $t_{mode}$, $Cond_{list}$, and $d_{max}$ are given to the algorithm as input. As a result, the first task will be performed using the IG threshold to eliminate patterns and discover interesting associations. After the first task is completed, the generated patterns are used as input for the second task, which removes irrelevant selectors from these patterns to obtain patterns with a large amount of information and dependencies on the target, while minimizing complexity. This is achieved by relying on IG and OR measures.

\subsubsection{Discovering relevant associations}

In order to select the selectors that surpass the IG $thre$ and contribute the most information to the problem, the first step, which is finding interesting associations, begins with the calculation of an IG threshold for each selector. Equation~\ref{eq:Thre} is used to compute the IG threshold:

\begin{equation}
thre = \sqrt{\frac{n\sum_{i=1}^{n}x_{i}^2 - (\sum_{i=1}^{n}x_{i})^2}{n(n-1)}}
\label{eq:Thre}
\end{equation}

Where the $n$ term indicates the total number of selectors that are contemplated, and the $x_i$ term indicates the IG value of a particular selector. As a result, among all of the possible selectors, the IG threshold for each subgroup at each exploration step is calculated.

\begin{algorithm}[H]
\caption{InfoGained SD Algorithm}\label{alg:igsd}
\KwData{Dataset $D$, maximum depth $dmax$, threshold mode $t\_mode$, condition attributes $CondList$}
M $\gets$ filterByThreshold(all selectors in $D$, t\_mode)\;
\For{$i\leftarrow 2$ \KwTo $dmax$}{
    M\_aux $\gets$ []\;
    \For{$j\leftarrow 0$ \KwTo len(M)}{
        cands $\gets$ get selector candidates which contain the attributes presented in $CondList$ ([Patterns with length=i with parent node $= M[j]$])\;
        final\_cands $\gets$ filterByThreshold(cands,t\_mode)\;
        M\_aux $\gets$ M\_aux + final\_cands\;
        }
    M $\gets$ M\_aux\;
    }
R $\gets$ []\;
\For{$k\leftarrow 0$ \KwTo len(M)}{
    s $\gets$ calculate\_optimal\_cut(M[k])\;
	R $\gets$ R + s\;
}
return R\;
\end{algorithm}


Algorithm~\ref{alg:igsd} shows the steps performed during the interesting association discovery phase. In line 1 of Algorithm~\ref{alg:igsd}, variable $M$ will contain subgroups with one selector, i.e., of length 1, with an IG value higher or equal to an IG threshold. Depending on the parameter $t_{mode}$, this IG threshold will be either the maximum IG value of all the subgroups with one selector ($t_{mode}$='maximum') or the value computed using (Equation~\ref{eq:Thre}) ($t_{mode}$='dynamic').  

Subgroups are constructed in an iterative process in lines 2 to 10, adding selectors with IG values equal to or greater than $thre$ at each step (Equation~\ref{eq:Thre}). Each pattern contained in $M$ obtained in line 1 will serve as the basis for this iterative process. Consequently, in line 5 for each pattern ($j$) in $M$, another selector is added to design $M[j]$ expanding the length by 1 up to an all-out design length of $i$. Furthermore, line 5 in Algorithm1 stores in $cands$ variable the new patterns of length $i$ that contain attributes specified in user-provided argument $Cond_{list}$, on the off chance that it isn't empty. Then, at that point, in line 6, patterns stored in $cands$ variable are filtered by IG value by computing dynamic threshold and according to argument $t_{mode}$, as in line 1 of Algorithm~\ref{alg:igsd}. This process of iteration will continue until the $d_{max}$ parameter is reached.

\begin{figure}[H]
\centering
\includegraphics[width=\columnwidth, height=4in]{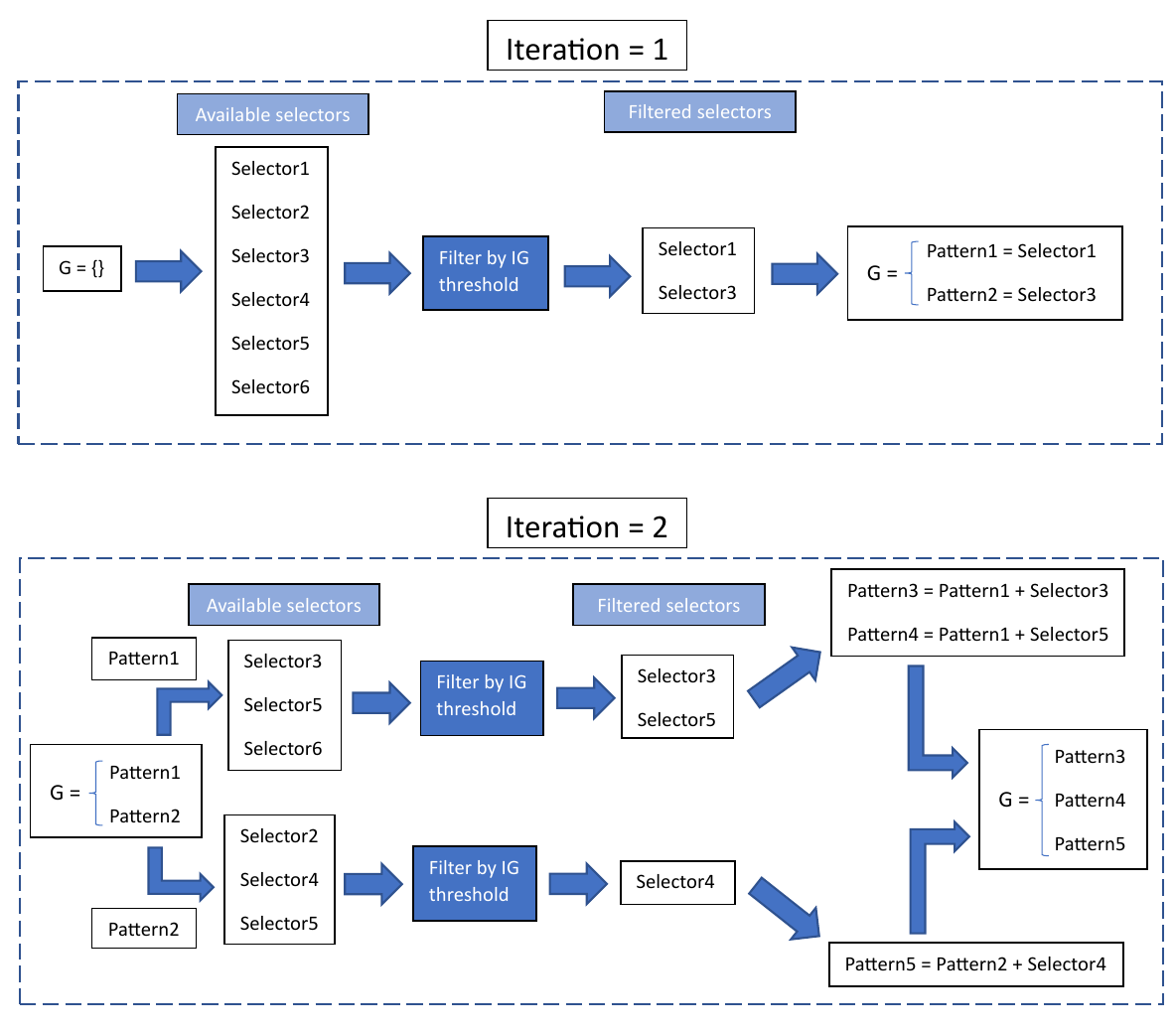}
\caption{Iterative optimization process for pattern search}
\label{algo_example}
\end{figure}

In addition, using a $d_{max}$ = 2 as an illustration, Fig. \ref{algo_example} provides a better understanding of how associations are constructed. As can be seen, in Iteration = 1 schema, an IG threshold is computed utilizing the IG values of available selectors from Selector1, to Selector6. Selectors 1 and 3 will be chosen to build the patterns (Patterns 1 and 2) in this first iteration because they exceed the threshold after the threshold was calculated. From here, the algorithm iterates for each pattern from the previous iteration (such as Pattern1 and Pattern2) in the second iteration. For the first pattern, Selectors 3, 5 and 6 are candidates since the combination of Pattern1 and these selectors, is present in the input data set. On the other hand, for the second pattern, Selector2, Selector4, and Selector5 are the possible selectors to add. It is important to notice that for each pattern of the iterative process, a different IG threshold is calculated for each one. So,  for Pattern1, only Selector3, and Selector5 surpass the particular threshold, so they will be added to Pattern1, getting Pattern3 and Pattern4, of length 2 every one. However, only Selector 4 surpasses the required threshold, so it is added to Pattern 2, resulting in Pattern 5.

\subsubsection{Removing irrelevant information}\mbox{}
\newline

\begin{figure}[H]%
    \centering
    \subfloat[\centering Complexity reduction process for patterns]{{\includegraphics[width=.45\columnwidth,height=2.5in]{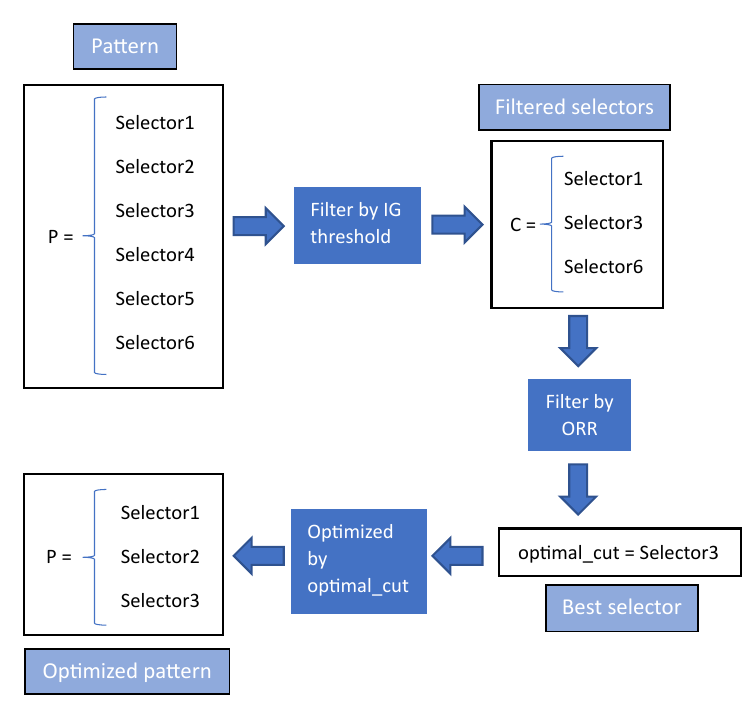} }}%
    \qquad
    \subfloat[\centering Pattern diagram example]{{\includegraphics[width=.45\columnwidth,height=2.5in]{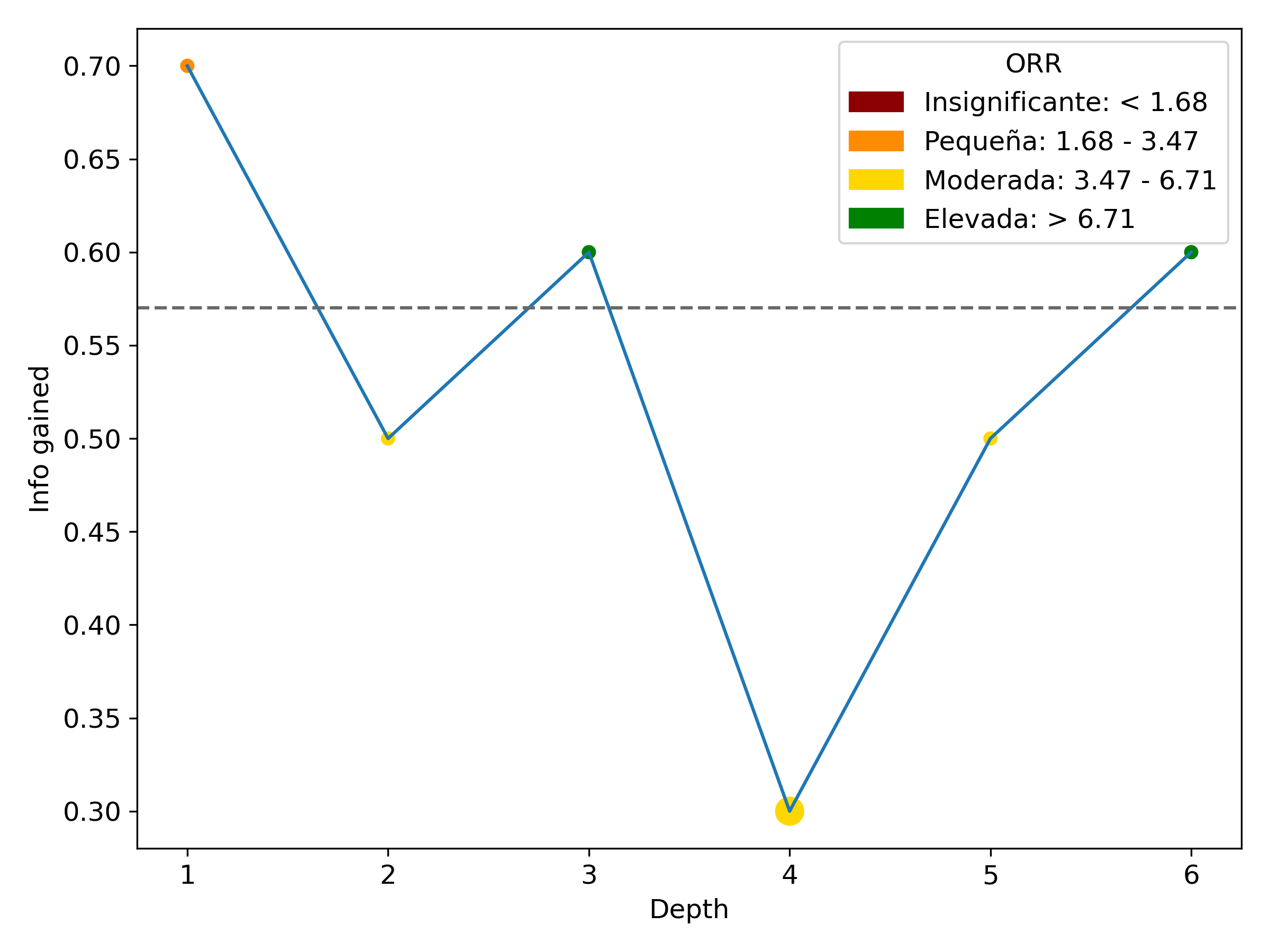} }}%
    \caption{Example of complexity pattern reduction}%
    \label{fig:alg2_example}%
\end{figure}

For the second task, pattern complexity is reduced by removing irrelevant information from patterns after they are generated in the first task. The purpose of this step is to determine which selectors of a pattern are not giving valuable or important information.
Thus, Figure~\ref{fig:alg2_example} shows that among the 6 selectors of a given pattern, selector 3 is identified as the best selector since its IG value is over the IG threshold (dashed line) and has a high ORR value. Based on the identified optimal selector, the pattern is cut and selectors 4, 5 and 6 are removed as irrelevant information.

This step initiates by setting of list $R$ for storing optimized patterns in Algorithm~\ref{alg:igsd}, line 11. As a result, on lines 11 and 14, IGSD stores the optimal cut in the $R$ output list for each pattern that was returned in the variable $M$ by the first step of Algorithm~\ref{alg:igsd}.

Algorithm~\ref{alg:cut} demonstrates the procedures used to determine a pattern's best cut point (i.e. selector),
First of all, Line 1 of Algorithm~\ref{alg:cut} converts $OR$ values to $ORR$, and Line 2 calculates an IG threshold using all the selectors presented in the input pattern in accordance with Equation \ref{eq:Thre}). Thusly, the calculation will dispose of those selectors with IG values lower than the IG threshold. Besides, line 3 filters not statistically relevant selectors by removing those ones with a p-value measure below 0.05.

After the filtering of selectors, if only one selector remains, its position will be used to cut the pattern and returned as the $optimal\_cut$ cut (lines 4 through 6 of Algorithm~\ref{alg:cut}). On the other hand, line 8 of Algorithm~\ref{alg:cut} will iterate over the candidates as follows in order to determine the best cut:
\begin{itemize}
    \item At the beginning, the first selector of potential candidates is considered as the $optimal\_cut$ ((Algorithm \ref{alg:cut}, line 7).
    \item The $optimal\_cut$ is not updated in lines 9 to 10 until a candidate selector's ORR improves the $optimal\_cut$ ORR.
    \item There are two conditions to stop the iteration in lines 12 to 14. Whether the up-and-comer selector in the ongoing iteration has a lower ORR than the $optimal\_cut$ ORR or the ongoing examined selector isn't sequential to the recently analyzed selector and its ORR is equivalent to the ORR $optimal\_cut$. In such cases, the iteration stops because new elements should be added to a pattern only when the $optimal\_cut$ ORR improves.
\end{itemize}

\begin{algorithm}[H]
\caption{calculate\_optimal\_cut Algorithm}\label{alg:cut}
\KwData{Information gained list ig, Odd ratio list or, p-value list pv, Pattern p}
ORR $\gets$ [values of Odd ratio list are transformed into ranges]\;
cut\_candidates $\gets$ filterByThreshold(ig,t\_mode='dynamic')\;
cut\_candidates $\gets$ [elements in cut\_candidates with p\_value $\leq$ 0.05]\;
\If{len(cut\_candidates) == 1}{
      return p[:ig.index(cut\_candidates[0])]\;
    }
optimal\_cut $\gets$ ORR[0]\;
\For{$i\leftarrow 1$ \KwTo len(ORR)}{
    \If{ORR[i] $>$ ORR[i-1]}{
        optimal\_cut $\gets$ ORR[i]\;
    }
    \ElseIf{(ORR[i] == ORR[i-1] $\land$ ORR[i] is not consecutive to ORR[i-1]]) $\lor$ (ORR[i] $<$ ORR[i-1])}{
        break\;
    }
}
return p[:ORR.index(optimal\_cut)]\;
\end{algorithm}


\subsubsection{Discovered patterns validation}

This section describes the validation performed of patterns provided by the compared algorithms. Thus, the pattern validation process is described as well as the inter-rater agreement indices used for assessing the validation.\\

To validate the patterns obtained through different SD algorithms, seven oncologists were recruited to assess and rate the medical relevance of patterns. All raters received the same data in the same order. The goal of this procedure was to understand whether these patterns were providing useful information to clinicians or not. Hence, two options could be chosen for each pattern:

\begin{itemize}
\item Accept: if the information contained therein is relevant or of interest, regardless of whether its content is in line with CG or clinical experience, or whether it is something new.

\item Reject: if the information provided does not add anything clinically relevant or does not contain sufficient information to be considered of interest. 

\end{itemize}

\textbf{Inter-rater agreement metrics}

In order to assess the inter-rater agreement of the pattern evaluation, AC1 index was used. Also called "first-order agreement coefficient”, it adjusts the overall probability based on the chance that raters may agree on a rating, despite the fact that one or all of them may have given a random value \cite{wongpakaran_comparison_2013}. It can be calculated as follows: 

\begin{equation}
AC1 = \frac{p-e(\gamma)}{1-e(\gamma)}
\label{eq:AC1}
\end{equation}
Where, 
\begin{equation}
e(\gamma) = 2q (1-q), \quad p = \frac{A+D}{N} \quad\mathrm{and}\quad q = \frac{A1+B1}{2N}
\end{equation}

Here, \textit{A} is the number of times both raters accept the patterns, \textit{D} is the number of times both raters reject the patterns and \textit{N} is the total sample size. Thus, \textit{p} is the proportion of observed agreement and \textit{e($\gamma$)} is the proportion of the expected agreement.

Furthermore, Intraclass correlation coefficient (ICC) was used to evaluate the inter-rater reliability. It can be stated as: 

\begin{equation}
ICC = \frac{\sigma_b^2}{\sigma_b^2-\sigma_w^2}
\label{eq:ICC}
\end{equation}

Here, $\sigma_b^2$ is the variance between subjects and $\sigma_w^2$ is the variance within subjects. p-value and confidence interval (CI) is also provided for each index.\\

%% file: tex/clinical_guidelines.tex

%% file: tex/results.tex
\section{Results}
In this section, we present the performance results of SD algorithms in 11 data sets. Also, for the P4Lucat dataset it was possible to perform validation by a group of problem domain experts, i.e. clinicians. Validation results are reported and compared with performance metrics obtained by each SD method for the P4Lucat dataset.

\subsection{Subgroup Discovery algorithms}
\label{sec:Results1}
Initially, performance comparison of FSSD, SSD++, and IGSD algorithms has been made based on descriptive and predictive measures. FSSD and SSD++ were used with default parameters, and the two versions of IGSD (i.e. IGSD-M and IGSD-T) were tested according to the two possible values of algorithm argument $t_{mode}$. On the other hand, argument $d_{max}$ was not manually set. Thus, the maximum exploration depth was, by default, the number of variables of the input dataset. Also, since FSSD and SSD++ define a timer of one hour to limit computations during pattern search, we have define a similar timer for both IGSD versions for comparison purposes. Additionally, for the P4Lucat dataset, results for both versions of IGSD were obtained by including cancer stage and first treatment variables in the $Cond_{list}$ argument. These variables, also included in a previous study \cite{CBMS22}, were identified by clinicians as essential and required to be present in discovered patterns.

\begin{table}[H]
  \setlength\tabcolsep{6pt}
  \caption{{Statistical comparison using SD algorithms for Numeric datasets.}}
  \label{tab:resNumDatasets}
  \hfill%
  \resizebox{\textwidth}{!}{%
  \subfloat[IRIS dataset]{%
    \label{tab:iris_stats}%
    \begin{tabular}{|ccccc|l}
\cline{1-5}
\multicolumn{5}{|c|}{\textbf{IRIS}} &
   \\ \cline{1-5}
\multicolumn{1}{|c|}{} &
  \multicolumn{1}{c|}{\textbf{IGSD-M}} &
  \multicolumn{1}{c|}{\textbf{IGSD-T}} &
  \multicolumn{1}{c|}{\textbf{FSSD}} &
  \textbf{SSD++} &
   \\ \cline{1-5}
\multicolumn{1}{|c|}{\textbf{size}} &
  \multicolumn{1}{c|}{\cellcolor[HTML]{63BE7B}9} &
  \multicolumn{1}{c|}{\cellcolor[HTML]{B1D78C}7} &
  \multicolumn{1}{c|}{\cellcolor[HTML]{FFEF9C}5} &
  \cellcolor[HTML]{FFEF9C}5 &
   \\ \cline{1-5}
\multicolumn{1}{|c|}{\textbf{length}} &
  \multicolumn{1}{c|}{\cellcolor[HTML]{FFEF9C}1.1} &
  \multicolumn{1}{c|}{\cellcolor[HTML]{FDEF9C}1.14} &
  \multicolumn{1}{c|}{\cellcolor[HTML]{63BE7B}4} &
  \cellcolor[HTML]{FAEE9B}1.2 &
   \\ \cline{1-5}
\multicolumn{1}{|c|}{\textbf{coverage}} &
  \multicolumn{1}{c|}{\cellcolor[HTML]{F0EB99}0.21} &
  \multicolumn{1}{c|}{\cellcolor[HTML]{FFEF9C}0.18} &
  \multicolumn{1}{c|}{\cellcolor[HTML]{EFEA99}0.212} &
  \cellcolor[HTML]{63BE7B}0.48 &
   \\ \cline{1-5}
\multicolumn{1}{|c|}{\textbf{wracc}} &
  \multicolumn{1}{c|}{\cellcolor[HTML]{B2D78C}0.12} &
  \multicolumn{1}{c|}{\cellcolor[HTML]{B2D78C}0.12} &
  \multicolumn{1}{c|}{\cellcolor[HTML]{63BE7B}0.13} &
  \cellcolor[HTML]{FFEF9C}0.11 &
   \\ \cline{1-5}
\multicolumn{1}{|c|}{\textbf{confidence}} &
  \multicolumn{1}{c|}{\cellcolor[HTML]{78C580}0.94} &
  \multicolumn{1}{c|}{\cellcolor[HTML]{63BE7B}0.98} &
  \multicolumn{1}{c|}{\cellcolor[HTML]{6EC27E}0.96} &
  \cellcolor[HTML]{FFEF9C}0.67 &
   \\ \cline{1-5}
\multicolumn{1}{|c|}{\textbf{accuracy}} &
  \multicolumn{1}{c|}{\cellcolor[HTML]{70C37E}0.84} &
  \multicolumn{1}{c|}{\cellcolor[HTML]{70C37E}0.84} &
  \multicolumn{1}{c|}{\cellcolor[HTML]{63BE7B}0.85} &
  \cellcolor[HTML]{FFEF9C}0.73 &
   \\ \cline{1-5}
\multicolumn{1}{|c|}{\textbf{info\_gained}} &
  \multicolumn{1}{c|}{\cellcolor[HTML]{EFEA99}0.327} &
  \multicolumn{1}{c|}{\cellcolor[HTML]{EFEA99}0.327} &
  \multicolumn{1}{c|}{\cellcolor[HTML]{63BE7B}0.47} &
  \cellcolor[HTML]{FFEF9C}0.31 &
   \\ \cline{1-5}
\multicolumn{1}{|c|}{\textbf{odd-range}} &
  \multicolumn{1}{c|}{\cellcolor[HTML]{63BE7B}4} &
  \multicolumn{1}{c|}{\cellcolor[HTML]{63BE7B}4} &
  \multicolumn{1}{c|}{\cellcolor[HTML]{63BE7B}4} &
  \cellcolor[HTML]{FFEF9C}3.4 &
   \\ \cline{1-5}
\multicolumn{1}{|c|}{\textbf{p-value}} &
  \multicolumn{1}{c|}{\cellcolor[HTML]{63BE7B}4.23E-10} &
  \multicolumn{1}{c|}{\cellcolor[HTML]{63BE7B}1.95E-10} &
  \multicolumn{1}{c|}{\cellcolor[HTML]{FFEF9C}6.20E-02} &
  \cellcolor[HTML]{78C47F}8.50E-03 &
   \\ \cline{1-5}
\end{tabular}}
  \hfill%
  \subfloat[ECHO dataset]{%
    \label{tab:echo_stats}%
    \begin{tabular}{|ccccc|l}
\cline{1-5}
\multicolumn{5}{|c|}{\textbf{ECHO}} &
   \\ \cline{1-5}
\multicolumn{1}{|c|}{} &
  \multicolumn{1}{c|}{\textbf{IGSD-M}} &
  \multicolumn{1}{c|}{\textbf{IGSD-T}} &
  \multicolumn{1}{c|}{\textbf{FSSD}} &
  \textbf{SSD++} &
   \\ \cline{1-5}
\multicolumn{1}{|c|}{\textbf{size}} &
  \multicolumn{1}{c|}{\cellcolor[HTML]{9ED188}9} &
  \multicolumn{1}{c|}{\cellcolor[HTML]{63BE7B}12} &
  \multicolumn{1}{c|}{\cellcolor[HTML]{FFEF9C}4} &
  - &
   \\ \cline{1-5}
\multicolumn{1}{|c|}{\textbf{length}} &
  \multicolumn{1}{c|}{\cellcolor[HTML]{FFEF9C}2} &
  \multicolumn{1}{c|}{\cellcolor[HTML]{FCEE9C}2.09} &
  \multicolumn{1}{c|}{\cellcolor[HTML]{63BE7B}6} &
  - &
   \\ \cline{1-5}
\multicolumn{1}{|c|}{\textbf{coverage}} &
  \multicolumn{1}{c|}{\cellcolor[HTML]{EFEA99}0.1} &
  \multicolumn{1}{c|}{\cellcolor[HTML]{FFEF9C}0.083} &
  \multicolumn{1}{c|}{\cellcolor[HTML]{63BE7B}0.247} &
  - &
   \\ \cline{1-5}
\multicolumn{1}{|c|}{\textbf{wracc}} &
  \multicolumn{1}{c|}{\cellcolor[HTML]{F7ED9B}0.03} &
  \multicolumn{1}{c|}{\cellcolor[HTML]{FFEF9C}0.028} &
  \multicolumn{1}{c|}{\cellcolor[HTML]{63BE7B}0.065} &
  - &
   \\ \cline{1-5}
\multicolumn{1}{|c|}{\textbf{confidence}} &
  \multicolumn{1}{c|}{\cellcolor[HTML]{63BE7B}0.91} &
  \multicolumn{1}{c|}{\cellcolor[HTML]{77C580}0.9} &
  \multicolumn{1}{c|}{\cellcolor[HTML]{FFEF9C}0.83} &
  - &
   \\ \cline{1-5}
\multicolumn{1}{|c|}{\textbf{accuracy}} &
  \multicolumn{1}{c|}{\cellcolor[HTML]{FFEF9C}0.61} &
  \multicolumn{1}{c|}{\cellcolor[HTML]{D8E394}0.65} &
  \multicolumn{1}{c|}{\cellcolor[HTML]{63BE7B}0.77} &
  - &
   \\ \cline{1-5}
\multicolumn{1}{|c|}{\textbf{info\_gained}} &
  \multicolumn{1}{c|}{\cellcolor[HTML]{FDEF9C}0.05} &
  \multicolumn{1}{c|}{\cellcolor[HTML]{FFEF9C}0.049} &
  \multicolumn{1}{c|}{\cellcolor[HTML]{63BE7B}0.117} &
  - &
   \\ \cline{1-5}
\multicolumn{1}{|c|}{\textbf{odd-range}} &
  \multicolumn{1}{c|}{\cellcolor[HTML]{FFEF9C}3.89} &
  \multicolumn{1}{c|}{\cellcolor[HTML]{D5E294}3.92} &
  \multicolumn{1}{c|}{\cellcolor[HTML]{63BE7B}4} &
  - &
   \\ \cline{1-5}
\multicolumn{1}{|c|}{\textbf{p-value}} &
  \multicolumn{1}{c|}{\cellcolor[HTML]{FFEF9C}1.40E-02} &
  \multicolumn{1}{c|}{\cellcolor[HTML]{EEE998}1.30E-02} &
  \multicolumn{1}{c|}{\cellcolor[HTML]{63BE7B}4.70E-03} &
  - &
   \\ \cline{1-5}
\end{tabular}}%
}
  \hspace{\fill}

  \vspace{12pt}
  
  \hfill%
  \resizebox{\textwidth}{!}{%
  \subfloat[HEART dataset]{%
    \label{tab:heart_stats}
    \begin{tabular}{|ccccc|l}
\cline{1-5}
\multicolumn{5}{|c|}{\textbf{HEART}} &
   \\ \cline{1-5}
\multicolumn{1}{|c|}{} &
  \multicolumn{1}{c|}{\textbf{IGSD-M}} &
  \multicolumn{1}{c|}{\textbf{IGSD-T}} &
  \multicolumn{1}{c|}{\textbf{FSSD}} &
  \textbf{SSD++} &
   \\ \cline{1-5}
\multicolumn{1}{|c|}{\textbf{size}} &
  \multicolumn{1}{c|}{\cellcolor[HTML]{D8E394}7} &
  \multicolumn{1}{c|}{\cellcolor[HTML]{63BE7B}22} &
  \multicolumn{1}{c|}{\cellcolor[HTML]{FFEF9C}2} &
  \cellcolor[HTML]{E0E696}6 &
   \\ \cline{1-5}
\multicolumn{1}{|c|}{\textbf{length}} &
  \multicolumn{1}{c|}{\cellcolor[HTML]{FFEF9C}1.86} &
  \multicolumn{1}{c|}{\cellcolor[HTML]{FBEE9C}2.14} &
  \multicolumn{1}{c|}{\cellcolor[HTML]{63BE7B}13} &
  \cellcolor[HTML]{FFEF9C}1.83 &
   \\ \cline{1-5}
\multicolumn{1}{|c|}{\textbf{coverage}} &
  \multicolumn{1}{c|}{\cellcolor[HTML]{9DD088}0.257} &
  \multicolumn{1}{c|}{\cellcolor[HTML]{DDE595}0.22} &
  \multicolumn{1}{c|}{\cellcolor[HTML]{FFEF9C}0.2} &
  \cellcolor[HTML]{63BE7B}0.29 &
   \\ \cline{1-5}
\multicolumn{1}{|c|}{\textbf{wracc}} &
  \multicolumn{1}{c|}{\cellcolor[HTML]{63BE7B}0.089} &
  \multicolumn{1}{c|}{\cellcolor[HTML]{87CA83}0.075} &
  \multicolumn{1}{c|}{\cellcolor[HTML]{93CE86}0.07} &
  \cellcolor[HTML]{FFEF9C}0.027 &
   \\ \cline{1-5}
\multicolumn{1}{|c|}{\textbf{confidence}} &
  \multicolumn{1}{c|}{\cellcolor[HTML]{63BE7B}0.87} &
  \multicolumn{1}{c|}{\cellcolor[HTML]{6BC17D}0.86} &
  \multicolumn{1}{c|}{\cellcolor[HTML]{89CA83}0.82} &
  \cellcolor[HTML]{FFEF9C}0.66 &
   \\ \cline{1-5}
\multicolumn{1}{|c|}{\textbf{accuracy}} &
  \multicolumn{1}{c|}{\cellcolor[HTML]{63BE7B}0.67} &
  \multicolumn{1}{c|}{\cellcolor[HTML]{8BCB84}0.64} &
  \multicolumn{1}{c|}{\cellcolor[HTML]{98CF87}0.63} &
  \cellcolor[HTML]{FFEF9C}0.55 &
   \\ \cline{1-5}
\multicolumn{1}{|c|}{\textbf{info\_gained}} &
  \multicolumn{1}{c|}{\cellcolor[HTML]{B1D78C}0.137} &
  \multicolumn{1}{c|}{\cellcolor[HTML]{EAE998}0.11} &
  \multicolumn{1}{c|}{\cellcolor[HTML]{FFEF9C}0.1} &
  \cellcolor[HTML]{63BE7B}0.174 &
   \\ \cline{1-5}
\multicolumn{1}{|c|}{\textbf{odd-range}} &
  \multicolumn{1}{c|}{\cellcolor[HTML]{63BE7B}4} &
  \multicolumn{1}{c|}{\cellcolor[HTML]{63BE7B}4} &
  \multicolumn{1}{c|}{\cellcolor[HTML]{B1D78C}3.5} &
  \cellcolor[HTML]{FFEF9C}3 &
   \\ \cline{1-5}
\multicolumn{1}{|c|}{\textbf{p-value}} &
  \multicolumn{1}{c|}{\cellcolor[HTML]{63BE7B}1.37E-09} &
  \multicolumn{1}{c|}{\cellcolor[HTML]{63BE7B}5.13E-07} &
  \multicolumn{1}{c|}{\cellcolor[HTML]{FFEF9C}2.30E-04} &
  \cellcolor[HTML]{63BE7B}2.62E-08 &
   \\ \cline{1-5}
\end{tabular}}%
  \hfill%
  \subfloat[MAGIC dataset]{%
    \label{tab:magic_stats}
    \begin{tabular}{|ccccc|l}
\cline{1-5}
\multicolumn{5}{|c|}{\textbf{MAGIC}} &
   \\ \cline{1-5}
\multicolumn{1}{|c|}{} &
  \multicolumn{1}{c|}{\textbf{IGSD-M}} &
  \multicolumn{1}{c|}{\textbf{IGSD-T}} &
  \multicolumn{1}{c|}{\textbf{FSSD}} &
  \textbf{SSD++} &
   \\ \cline{1-5}
\multicolumn{1}{|c|}{\textbf{size}} &
  \multicolumn{1}{c|}{\cellcolor[HTML]{FFEF9C}9} &
  \multicolumn{1}{c|}{\cellcolor[HTML]{B1D78C}51} &
  \multicolumn{1}{c|}{-} &
  \cellcolor[HTML]{63BE7B}92 &
   \\ \cline{1-5}
\multicolumn{1}{|c|}{\textbf{length}} &
  \multicolumn{1}{c|}{\cellcolor[HTML]{FFEF9C}3.56} &
  \multicolumn{1}{c|}{\cellcolor[HTML]{F4EC9A}3.59} &
  \multicolumn{1}{c|}{-} &
  \cellcolor[HTML]{63BE7B}3.97 &
   \\ \cline{1-5}
\multicolumn{1}{|c|}{\textbf{coverage}} &
  \multicolumn{1}{c|}{\cellcolor[HTML]{63BE7B}0.042} &
  \multicolumn{1}{c|}{\cellcolor[HTML]{92CD85}0.039} &
  \multicolumn{1}{c|}{-} &
  \cellcolor[HTML]{FFEF9C}0.032 &
   \\ \cline{1-5}
\multicolumn{1}{|c|}{\textbf{wracc}} &
  \multicolumn{1}{c|}{\cellcolor[HTML]{63BE7B}0.02} &
  \multicolumn{1}{c|}{\cellcolor[HTML]{6EC27E}0.019} &
  \multicolumn{1}{c|}{-} &
  \cellcolor[HTML]{FFEF9C}0.0048 &
   \\ \cline{1-5}
\multicolumn{1}{|c|}{\textbf{confidence}} &
  \multicolumn{1}{c|}{\cellcolor[HTML]{63BE7B}0.96} &
  \multicolumn{1}{c|}{\cellcolor[HTML]{6BC17D}0.95} &
  \multicolumn{1}{c|}{-} &
  \cellcolor[HTML]{FFEF9C}0.74 &
   \\ \cline{1-5}
\multicolumn{1}{|c|}{\textbf{accuracy}} &
  \multicolumn{1}{c|}{\cellcolor[HTML]{63BE7B}0.59} &
  \multicolumn{1}{c|}{\cellcolor[HTML]{63BE7B}0.59} &
  \multicolumn{1}{c|}{-} &
  \cellcolor[HTML]{FFEF9C}0.52 &
   \\ \cline{1-5}
\multicolumn{1}{|c|}{\textbf{info\_gained}} &
  \multicolumn{1}{c|}{\cellcolor[HTML]{63BE7B}0.036} &
  \multicolumn{1}{c|}{\cellcolor[HTML]{7AC580}0.034} &
  \multicolumn{1}{c|}{-} &
  \cellcolor[HTML]{FFEF9C}0.022 &
   \\ \cline{1-5}
\multicolumn{1}{|c|}{\textbf{odd-range}} &
  \multicolumn{1}{c|}{\cellcolor[HTML]{63BE7B}4} &
  \multicolumn{1}{c|}{\cellcolor[HTML]{67C07C}3.98} &
  \multicolumn{1}{c|}{-} &
  \cellcolor[HTML]{FFEF9C}3.16 &
   \\ \cline{1-5}
\multicolumn{1}{|c|}{\textbf{p-value}} &
  \multicolumn{1}{c|}{\cellcolor[HTML]{63BE7B}2.01E-84} &
  \multicolumn{1}{c|}{\cellcolor[HTML]{63BE7B}7.83E-46} &
  \multicolumn{1}{c|}{-} &
  \cellcolor[HTML]{FFEF9C}8.73E-11 &
   \\ \cline{1-5}
\end{tabular}}%
}
  \hspace{\fill}
\end{table}

Table~\ref{tab:resNumDatasets} shows the obtained metric values for the numeric datasets considered. Also, table~\ref{tab:resNumDatasets} uses a color range (from green to yellow) to indicate the best and worst metric value for each row, being thick green the best and thick yellow the worst. The summary of results in terms of descriptive and predictive measures is as follows, considering that SSD++ algorithm was not able to discover patterns in the ECHO dataset as well as FSSD algorithm for the MAGIC dataset:
\begin{itemize}
\item Regarding the number of patterns found (i.e. size), both versions of IGSD obtain a higher number of patterns with respect to FSSD and SSD++, except for MAGIG dataset where SSD++ provides 92 patterns, versus 9 provided by IGSD-M and 51 provided by IGSD-T.
\item In terms of pattern complexity of the rule set (i.e. length), FSSD produces the set of patterns with the largest number of variables. On the other hand, SSD++ produces patterns of similar complexity than IGSD-M and IGSD-T. However, both IGSD versions are the only ones capable of discovering patterns for all datasets.
\item Regarding the average coverage per rule, considering the IRIS and HEART dataset, patterns produced by SSD++ have higher values (0.48, 0.29) than the patterns produced by both IGSD versions and FSSD algorithms with values similar to 0.2. Moreover, in the ECHO dataset, FSSD produced patterns with higher values (0.247) with respect to IGSD-M and IGSD-T (0.1, 0.083). On the other hand, concerning MAGIC dataset, IGSD-M and IGSD-T reports higher values (0.042, 0.039) respecting SDD++ (0.032).
\item Assessing the unusualness of the patterns (i.e. WRacc), all the algorithms, for the IRIS dataset, produce sets of patterns with a value similar to 0.12. Besides, in the ECHO dataset, FSSD produces a set of patterns with a higher value (0.247) than both IGSD versions (0.1, 0.083). On the other hand, concerning HEART and MAGIC datasets, IGSD-M produces slightly higher values (0.089, 0.02) than IGSD-T (0.075, 0.019), and much higher than SDD++ (0.027, 0.0048), respectively.

\item When evaluating the rule confidence, regarding ECHO and MAGIC datasets, the sets of patterns produced by IGSD-M (0.91, 0.96) and IGSD-T (0.9, 0.95), have higher confidence values than FSSD (0.83, -) and SSD++(-, 0.74). Furthermore, regarding IRIS dataset, IGSD-T produces a set of patterns with slightly more confidence value (0.98) than IGSD-M and FSSD (0.94, 0.96, respectively) and much higher than SSD++ (0.67). Moreover, concerning HEART dataset, IGSD-M and IGSD-T set of patterns produces higher values (0.87, 0.86) than FSSD (0.82) and SDD++ (0.66).

\item In terms of accuracy prediction, concerning the IRIS dataset, both IGSD versions and FSSD algorithms might be considered reliable models due to the accuracy values reported, 0.84, 0.84 and 0.85 respectively. Meanwhile SSD++ accuracy decreases to a value of 0.73.
Regarding the ECHO dataset, FSSD reports an accuracy value of 0.77, meanwhile, IGSD algorithm reports a lower accuracy value (0.61, 0.65 respectively). On the other hand, for HEART and MAGIC datasets, IGSD-M and IGSD-T report a slightly higher accuracy value (0.67, 0.59 and 0.64, 0.59, respectively) than FSSD (0.63, -) and SSD++ (0.55, 0.52) algorithms. 

\item Considering the average IG per rule, concerning the IRIS and ECHO datasets, IGSD-M and IGSD-T report patterns with less information gained (0.327, 0.05) and (0.327, 0.049) respectively than FSSD algorithm, with values of 0.47 and 0.117 respectively. Regarding the HEART dataset, SSD++ reports the highest information gained value (0.174) while IGSD-M, IGSD-T and FSSD report lower information gained values (0.137, 0.11 and 0.1), respectively. Furthermore, for MAGIC dataset, IGSD-M and IGSD-T produce sets of patterns with information gained values of 0.036 and 0.034 respectively while SSD++ produces a set of patterns with a lower information gained value of 0.022.
\item Taking into consideration the ORR metric, concerning the IRIS dataset, patterns produced by both IGSD versions and FSSD report an ORR of 4, meanwhile, SSD++ patterns produce a lower ORR of 3.4. Regarding the ECHO dataset, FSSD patterns report an ORR value of 4 while both IGSD versions report lower ORR values (3.89, 3.92), respectively. On the other hand, for HEART and MAGIC datasets, IGSD-M and IGSD-T set of patterns report the highest ORR values (4, 4) and (4, 3.98) respectively, while FSSD and SDD++ patterns report lower ORR values (3.5, -) and (3, 3.16) respectively.

\item In terms of p-value, regarding the IRIS dataset, both IGSD versions and SDD++ algorithms produce statistically significant patterns due to reporting a p-value below 0.05. Nevertheless, the IGSD algorithm reports a value below 0.001, which indicates that patterns might be considered more statistically significant. On the other hand, the FSSD algorithm reports patterns with a p-value above 0.05. In regards to the ECHO dataset, both IGSD versions and FSSD algorithms produce statistically significant patterns due to reporting a p-value below 0.05, nonetheless, FSSD reports a p-value below 0.01, being able to consider the patterns more statistically significant. Furthermore, concerning the HEART and MAGIC datasets, all the algorithms report a p-value below 0.001, which indicates a strong statistical significance.
\end{itemize}


\begin{table}[H]
\centering
  \setlength\tabcolsep{6pt}
  \caption{{Statistical comparison using SD algorithms for Nominal datasets.}}
  \label{tab:resNomDatasets}
  \hfill%
  \resizebox{\textwidth}{!}{%
  \subfloat[TIC-TAC-TOE dataset]{%
    \label{tab:tic_stats}%
    \begin{tabular}{|ccccc|l}
\cline{1-5}
\multicolumn{5}{|c|}{\textbf{TIC-TAC-TOE}} &
   \\ \cline{1-5}
\multicolumn{1}{|c|}{} &
  \multicolumn{1}{c|}{\textbf{IGSD-M}} &
  \multicolumn{1}{c|}{\textbf{IGSD-T}} &
  \multicolumn{1}{c|}{\textbf{FSSD}} &
  \textbf{SSD++} &
   \\ \cline{1-5}
\multicolumn{1}{|c|}{\textbf{size}} &
  \multicolumn{1}{c|}{\cellcolor[HTML]{FFEF9C}4} &
  \multicolumn{1}{c|}{\cellcolor[HTML]{FFEF9C}4} &
  \multicolumn{1}{c|}{\cellcolor[HTML]{B7D98D}10} &
  \cellcolor[HTML]{63BE7B}17 &
   \\ \cline{1-5}
\multicolumn{1}{|c|}{\textbf{length}} &
  \multicolumn{1}{c|}{\cellcolor[HTML]{63BE7B}3} &
  \multicolumn{1}{c|}{\cellcolor[HTML]{63BE7B}3} &
  \multicolumn{1}{c|}{\cellcolor[HTML]{BBDA8E}2.6} &
  \cellcolor[HTML]{FFEF9C}2.29 &
   \\ \cline{1-5}
\multicolumn{1}{|c|}{\textbf{coverage}} &
  \multicolumn{1}{c|}{\cellcolor[HTML]{FFEF9C}0.073} &
  \multicolumn{1}{c|}{\cellcolor[HTML]{FFEF9C}0.073} &
  \multicolumn{1}{c|}{\cellcolor[HTML]{63BE7B}0.13} &
  \cellcolor[HTML]{8ACB84}0.116 &
   \\ \cline{1-5}
\multicolumn{1}{|c|}{\textbf{wracc}} &
  \multicolumn{1}{c|}{\cellcolor[HTML]{78C580}0.033} &
  \multicolumn{1}{c|}{\cellcolor[HTML]{78C580}0.033} &
  \multicolumn{1}{c|}{\cellcolor[HTML]{63BE7B}0.035} &
  \cellcolor[HTML]{FFEF9C}0.02 &
   \\ \cline{1-5}
\multicolumn{1}{|c|}{\textbf{confidence}} &
  \multicolumn{1}{c|}{\cellcolor[HTML]{63BE7B}1} &
  \multicolumn{1}{c|}{\cellcolor[HTML]{63BE7B}1} &
  \multicolumn{1}{c|}{\cellcolor[HTML]{C7DE90}0.93} &
  \cellcolor[HTML]{FFEF9C}0.89 &
   \\ \cline{1-5}
\multicolumn{1}{|c|}{\textbf{accuracy}} &
  \multicolumn{1}{c|}{\cellcolor[HTML]{B2D78C}0.57} &
  \multicolumn{1}{c|}{\cellcolor[HTML]{B2D78C}0.57} &
  \multicolumn{1}{c|}{\cellcolor[HTML]{63BE7B}0.58} &
  \cellcolor[HTML]{FFEF9C}0.56 &
   \\ \cline{1-5}
\multicolumn{1}{|c|}{\textbf{info\_gained}} &
  \multicolumn{1}{c|}{\cellcolor[HTML]{63BE7B}0.072} &
  \multicolumn{1}{c|}{\cellcolor[HTML]{63BE7B}0.072} &
  \multicolumn{1}{c|}{\cellcolor[HTML]{A5D389}0.059} &
  \cellcolor[HTML]{FFEF9C}0.041 &
   \\ \cline{1-5}
\multicolumn{1}{|c|}{\textbf{odd-range}} &
  \multicolumn{1}{c|}{\cellcolor[HTML]{63BE7B}4} &
  \multicolumn{1}{c|}{\cellcolor[HTML]{63BE7B}4} &
  \multicolumn{1}{c|}{\cellcolor[HTML]{81C882}3.8} &
  \cellcolor[HTML]{FFEF9C}2.94 &
   \\ \cline{1-5}
\multicolumn{1}{|c|}{\textbf{p-value}} &
  \multicolumn{1}{c|}{\cellcolor[HTML]{63BE7B}1.93E-13} &
  \multicolumn{1}{c|}{\cellcolor[HTML]{63BE7B}1.96E-13} &
  \multicolumn{1}{c|}{\cellcolor[HTML]{63BE7B}7.74E-12} &
  \cellcolor[HTML]{FFEF9C}6.30E-02 &
   \\ \cline{1-5}
\end{tabular}}
  \hfill%
  \subfloat[VOTE dataset]{%
    \label{tab:vote_stats}%
    \begin{tabular}{|ccccc|l}
\cline{1-5}
\multicolumn{5}{|c|}{\textbf{VOTE}} &
   \\ \cline{1-5}
\multicolumn{1}{|c|}{} &
  \multicolumn{1}{c|}{\textbf{IGSD-M}} &
  \multicolumn{1}{c|}{\textbf{IGSD-T}} &
  \multicolumn{1}{c|}{\textbf{FSSD}} &
  \textbf{SSD++} &
   \\ \cline{1-5}
\multicolumn{1}{|c|}{\textbf{size}} &
  \multicolumn{1}{c|}{\cellcolor[HTML]{ADD68B}13} &
  \multicolumn{1}{c|}{\cellcolor[HTML]{63BE7B}21} &
  \multicolumn{1}{c|}{\cellcolor[HTML]{C8DE91}10} &
  \cellcolor[HTML]{FFEF9C}4 &
   \\ \cline{1-5}
\multicolumn{1}{|c|}{\textbf{length}} &
  \multicolumn{1}{c|}{\cellcolor[HTML]{FFEF9C}1.46} &
  \multicolumn{1}{c|}{\cellcolor[HTML]{EEEA99}2.14} &
  \multicolumn{1}{c|}{\cellcolor[HTML]{63BE7B}7.7} &
  \cellcolor[HTML]{E5E797}2.5 &
   \\ \cline{1-5}
\multicolumn{1}{|c|}{\textbf{coverage}} &
  \multicolumn{1}{c|}{\cellcolor[HTML]{63BE7B}0.41} &
  \multicolumn{1}{c|}{\cellcolor[HTML]{79C580}0.37} &
  \multicolumn{1}{c|}{\cellcolor[HTML]{FFEF9C}0.12} &
  \cellcolor[HTML]{A2D289}0.293 &
   \\ \cline{1-5}
\multicolumn{1}{|c|}{\textbf{wracc}} &
  \multicolumn{1}{c|}{\cellcolor[HTML]{63BE7B}0.173} &
  \multicolumn{1}{c|}{\cellcolor[HTML]{81C882}0.15} &
  \multicolumn{1}{c|}{\cellcolor[HTML]{FFEF9C}0.05} &
  \cellcolor[HTML]{D3E293}0.085 &
   \\ \cline{1-5}
\multicolumn{1}{|c|}{\textbf{confidence}} &
  \multicolumn{1}{c|}{\cellcolor[HTML]{93CD86}0.91} &
  \multicolumn{1}{c|}{\cellcolor[HTML]{9AD087}0.9} &
  \multicolumn{1}{c|}{\cellcolor[HTML]{63BE7B}0.98} &
  \cellcolor[HTML]{FFEF9C}0.75 &
   \\ \cline{1-5}
\multicolumn{1}{|c|}{\textbf{accuracy}} &
  \multicolumn{1}{c|}{\cellcolor[HTML]{63BE7B}0.87} &
  \multicolumn{1}{c|}{\cellcolor[HTML]{87CA83}0.81} &
  \multicolumn{1}{c|}{\cellcolor[HTML]{FFEF9C}0.61} &
  \cellcolor[HTML]{D5E294}0.68 &
   \\ \cline{1-5}
\multicolumn{1}{|c|}{\textbf{info\_gained}} &
  \multicolumn{1}{c|}{\cellcolor[HTML]{63BE7B}0.45} &
  \multicolumn{1}{c|}{\cellcolor[HTML]{A4D389}0.33} &
  \multicolumn{1}{c|}{\cellcolor[HTML]{FFEF9C}0.16} &
  \cellcolor[HTML]{79C580}0.41 &
   \\ \cline{1-5}
\multicolumn{1}{|c|}{\textbf{odd-range}} &
  \multicolumn{1}{c|}{\cellcolor[HTML]{63BE7B}4} &
  \multicolumn{1}{c|}{\cellcolor[HTML]{63BE7B}4} &
  \multicolumn{1}{c|}{\cellcolor[HTML]{63BE7B}4} &
  \cellcolor[HTML]{FFEF9C}3.25 &
   \\ \cline{1-5}
\multicolumn{1}{|c|}{\textbf{p-value}} &
  \multicolumn{1}{c|}{\cellcolor[HTML]{63BE7B}8.35E-28} &
  \multicolumn{1}{c|}{\cellcolor[HTML]{63BE7B}5.24E-29} &
  \multicolumn{1}{c|}{\cellcolor[HTML]{FFEF9C}1.10E-01} &
  \cellcolor[HTML]{63BE7B}1.95E-39 &
   \\ \cline{1-5}
\end{tabular}}%
}
  \hspace{\fill}

  \vspace{4pt}
  
  \hfill%
  
  \resizebox{\textwidth}{!}{%
  \subfloat[P4LUCAT dataset]{%
    \label{tab:p4lucat_stats}
    
    \begin{tabular}{|ccccc|l}
\cline{1-5}
\multicolumn{5}{|c|}{\textbf{P4LUCAT}} &
   \\ \cline{1-5}
\multicolumn{1}{|c|}{} &
  \multicolumn{1}{c|}{\textbf{IGSD-M}} &
  \multicolumn{1}{c|}{\textbf{IGSD-T}} &
  \multicolumn{1}{c|}{\textbf{FSSD}} &
  \textbf{SSD++} &
   \\ \cline{1-5}
\multicolumn{1}{|c|}{\textbf{size}} &
  \multicolumn{1}{c|}{\cellcolor[HTML]{F2EB9A}19} &
  \multicolumn{1}{c|}{\cellcolor[HTML]{63BE7B}52} &
  \multicolumn{1}{c|}{\cellcolor[HTML]{EEEA99}20} &
  \cellcolor[HTML]{FFEF9C}16 &
   \\ \cline{1-5}
\multicolumn{1}{|c|}{\textbf{length}} &
  \multicolumn{1}{c|}{\cellcolor[HTML]{90CD85}3.1} &
  \multicolumn{1}{c|}{\cellcolor[HTML]{63BE7B}3.9} &
  \multicolumn{1}{c|}{\cellcolor[HTML]{9ED188}2.85} &
  \cellcolor[HTML]{FFEF9C}1.11 &
   \\ \cline{1-5}
\multicolumn{1}{|c|}{\textbf{coverage}} &
  \multicolumn{1}{c|}{\cellcolor[HTML]{FCEE9C}0.024} &
  \multicolumn{1}{c|}{\cellcolor[HTML]{FFEF9C}0.016} &
  \multicolumn{1}{c|}{\cellcolor[HTML]{DBE495}0.094} &
  \cellcolor[HTML]{63BE7B}0.35 &
   \\ \cline{1-5}
\multicolumn{1}{|c|}{\textbf{wracc}} &
  \multicolumn{1}{c|}{\cellcolor[HTML]{D6E394}0.0095} &
  \multicolumn{1}{c|}{\cellcolor[HTML]{FFEF9C}0.0068} &
  \multicolumn{1}{c|}{\cellcolor[HTML]{63BE7B}0.017} &
  \cellcolor[HTML]{BFDB8F}0.011 &
   \\ \cline{1-5}
\multicolumn{1}{|c|}{\textbf{confidence}} &
  \multicolumn{1}{c|}{\cellcolor[HTML]{63BE7B}0.83} &
  \multicolumn{1}{c|}{\cellcolor[HTML]{69C07D}0.81} &
  \multicolumn{1}{c|}{\cellcolor[HTML]{C3DD90}0.51} &
  \cellcolor[HTML]{FFEF9C}0.31 &
   \\ \cline{1-5}
\multicolumn{1}{|c|}{\textbf{accuracy}} &
  \multicolumn{1}{c|}{\cellcolor[HTML]{63BE7B}0.79} &
  \multicolumn{1}{c|}{\cellcolor[HTML]{92CD85}0.73} &
  \multicolumn{1}{c|}{\cellcolor[HTML]{8BCB84}0.74} &
  \cellcolor[HTML]{FFEF9C}0.59 &
   \\ \cline{1-5}
\multicolumn{1}{|c|}{\textbf{info\_gained}} &
  \multicolumn{1}{c|}{\cellcolor[HTML]{DEE595}0.016} &
  \multicolumn{1}{c|}{\cellcolor[HTML]{FFEF9C}0.011} &
  \multicolumn{1}{c|}{\cellcolor[HTML]{C9DE91}0.019} &
  \cellcolor[HTML]{63BE7B}0.034 &
   \\ \cline{1-5}
\multicolumn{1}{|c|}{\textbf{odd-range}} &
  \multicolumn{1}{c|}{\cellcolor[HTML]{66BF7C}3.84} &
  \multicolumn{1}{c|}{\cellcolor[HTML]{63BE7B}3.88} &
  \multicolumn{1}{c|}{\cellcolor[HTML]{B0D78C}2.8} &
  \cellcolor[HTML]{FFEF9C}1.69 &
   \\ \cline{1-5}
\multicolumn{1}{|c|}{\textbf{p-value}} &
  \multicolumn{1}{c|}{\cellcolor[HTML]{63BE7B}1.55E-03} &
  \multicolumn{1}{c|}{\cellcolor[HTML]{69C07C}7.97E-03} &
  \multicolumn{1}{c|}{\cellcolor[HTML]{84C881}3.30E-02} &
  \cellcolor[HTML]{FFEF9C}1.50E-01 &
   \\ \cline{1-5}
\end{tabular}}%

\hfill%
  \subfloat[GENBASE dataset]{%
    \label{tab:genbase_stats}
    \begin{tabular}{|ccccc|l}
\cline{1-5}
\multicolumn{5}{|c|}{\textbf{GENBASE}} &
   \\ \cline{1-5}
\multicolumn{1}{|c|}{} &
  \multicolumn{1}{c|}{\textbf{IGSD-M}} &
  \multicolumn{1}{c|}{\textbf{IGSD-T}} &
  \multicolumn{1}{c|}{\textbf{FSSD}} &
  \textbf{SSD++} &
   \\ \cline{1-5}
\multicolumn{1}{|c|}{\textbf{size}} &
  \multicolumn{1}{c|}{\cellcolor[HTML]{81C882}37041} &
  \multicolumn{1}{c|}{\cellcolor[HTML]{63BE7B}45625} &
  \multicolumn{1}{c|}{\cellcolor[HTML]{FFEF9C}32} &
  \cellcolor[HTML]{FFEF9C}33 &
   \\ \cline{1-5}
\multicolumn{1}{|c|}{\textbf{length}} &
  \multicolumn{1}{c|}{\cellcolor[HTML]{FFEF9C}2} &
  \multicolumn{1}{c|}{\cellcolor[HTML]{FFEF9C}2} &
  \multicolumn{1}{c|}{\cellcolor[HTML]{63BE7B}1151.9} &
  \cellcolor[HTML]{FFEF9C}1 &
   \\ \cline{1-5}
\multicolumn{1}{|c|}{\textbf{coverage}} &
  \multicolumn{1}{c|}{\cellcolor[HTML]{FFEF9C}0.0086} &
  \multicolumn{1}{c|}{\cellcolor[HTML]{E7E897}0.027} &
  \multicolumn{1}{c|}{\cellcolor[HTML]{E1E696}0.032} &
  \cellcolor[HTML]{63BE7B}0.128 &
   \\ \cline{1-5}
\multicolumn{1}{|c|}{\textbf{wracc}} &
  \multicolumn{1}{c|}{\cellcolor[HTML]{FFEF9C}0.008} &
  \multicolumn{1}{c|}{\cellcolor[HTML]{9DD188}0.02} &
  \multicolumn{1}{c|}{\cellcolor[HTML]{63BE7B}0.027} &
  \cellcolor[HTML]{9DD188}0.02 &
   \\ \cline{1-5}
\multicolumn{1}{|c|}{\textbf{confidence}} &
  \multicolumn{1}{c|}{\cellcolor[HTML]{63BE7B}0.99} &
  \multicolumn{1}{c|}{\cellcolor[HTML]{6DC27E}0.95} &
  \multicolumn{1}{c|}{\cellcolor[HTML]{6BC17D}0.96} &
  \cellcolor[HTML]{FFEF9C}0.348 &
   \\ \cline{1-5}
\multicolumn{1}{|c|}{\textbf{accuracy}} &
  \multicolumn{1}{c|}{\cellcolor[HTML]{75C47F}0.98} &
  \multicolumn{1}{c|}{\cellcolor[HTML]{75C47F}0.98} &
  \multicolumn{1}{c|}{\cellcolor[HTML]{63BE7B}0.99} &
  \cellcolor[HTML]{FFEF9C}0.9 &
   \\ \cline{1-5}
\multicolumn{1}{|c|}{\textbf{info\_gained}} &
  \multicolumn{1}{c|}{\cellcolor[HTML]{FFEF9C}0.048} &
  \multicolumn{1}{c|}{\cellcolor[HTML]{B0D78C}0.1} &
  \multicolumn{1}{c|}{\cellcolor[HTML]{63BE7B}0.15} &
  \cellcolor[HTML]{B0D78C}0.1 &
   \\ \cline{1-5}
\multicolumn{1}{|c|}{\textbf{odd-range}} &
  \multicolumn{1}{c|}{\cellcolor[HTML]{63BE7B}4} &
  \multicolumn{1}{c|}{\cellcolor[HTML]{63BE7B}4} &
  \multicolumn{1}{c|}{\cellcolor[HTML]{63BE7B}4} &
  \cellcolor[HTML]{63BE7B}4 &
   \\ \cline{1-5}
\multicolumn{1}{|c|}{\textbf{p-value}} &
  \multicolumn{1}{c|}{\cellcolor[HTML]{63BE7B}6.44E-08} &
  \multicolumn{1}{c|}{\cellcolor[HTML]{63BE7B}5.23E-08} &
  \multicolumn{1}{c|}{\cellcolor[HTML]{63BE7B}2.85E-51} &
  \cellcolor[HTML]{FFEF9C}2.60E-02 &
   \\ \cline{1-5}
\end{tabular}}%
}
  \hspace{\fill}
\end{table}

Table~\ref{tab:resNomDatasets} shows the obtained metric values for the nominal datasets considered. Also, table~\ref{tab:resNomDatasets} uses a color range (from green to yellow) to indicate the best and worst metric value for each row, being thick green the best and thick yellow the worst. The summary of results in terms of descriptive and predictive measures is as follows:
\begin{itemize}
\item Regarding the TIC-TAC-TOE dataset, SSD++ returns a higher number of discovered patterns (17) with respect to FSSD (10) and both IGSD versions (4). On the other hand, concerning the VOTE and P4Lucat dataset, IGSD-T was able to discover a higher amount of patterns (21, 52) in contradistinction to IGSD-M (13, 19) and FSSD (10, 20) and SSD++ (4, 16). Moreover, considering the GENBASE dataset, both IGSD versions produce a huge amount of patterns (37041, 45625), as opposed to FSSD and SSD++ algorithms (32, 33), respectively.

\item In terms of the complexity of the rule set (i.e. length), regarding the TIC-TAC-TOE and P4Lucat, both IGSD versions produce sets of patterns with the largest number of variables (3, 3.1) and (3, 3.9). In turn, SSD++ sets of patterns have the least complexity (2.29, 1.11), thus containing much less information. Furthermore, respecting VOTE and GENBASE datasets, FSSD patterns contain a much higher number of variables (7.7, 1151) than IGSD-M (1.46, 2), IGSD-T (2.14, 2) and SSD++ (2.5, 1), respectively.

\item Regarding the average coverage per rule, considering the TIC-TAC-TOE dataset, patterns produced by FSSD have the highest coverage value (0.13), while both IGSD versions sets of patterns have less coverage value (0.073). Furthermore, considering the VOTE dataset, IGSD-M reports a slightly higher value (0.41) than IGSD-T (0.37), meanwhile, FSSD returns a set of patterns with much less coverage (0.12). Finally, with respect to P4Lucat and GENBASE datasets, SSD++ produces a set of patterns with a high coverage value (0.35, 0.128), meanwhile, FSSD and both IGSD versions report a much lower coverage value, (0.094, 0.032) and (0.024, 0.0086 and 0.016, 0.027) respectively.

\item Assessing the unusualness of the patterns (i.e. WRacc), regarding the TIC-TAC-TOE dataset, FSSD and both IGSD versions produce sets of patterns with high unusualness values similar to 0.035. However, SSD++ set of patterns has lower unusualness value (0.02). Moreover, concerning the VOTE dataset, IGSD-M and IGSD-T report higher values (0.173, 0.15) than FSSD and SSD++ (0.05, 0.085), respectively. Finally, respecting P4Lucat and GENBASE datasets, FFSD produces patterns with higher unusualness value (0.017, 0.027) than IGSD-M (0.0095, 0.008), IGSD-T (0.0068, 0.02) and SSD++ (0.011, 0.02).

\item When evaluating patterns confidence, concerning the TIC-TAC-TOE and P4Lucat datasets, IGSD-M and IGSD-T report higher confidence values (1, 0.83) and (1, 0.81), respectively. However, in P4Lucat dataset, FSSD and SSD++ report much lower confidence values (0.51, 0.31). On the other hand, the set of patterns discovered by FSSD for the VOTE dataset reports a confidence value of 0.98, while, IGSD-M and IGSD-T confidence values are slightly lower (0.91, 0.9) and SSD++ confidence value is significantly lower (0.75). Finally, for GENBASE dataset, both IGSD versions and FSSD algorithms report high confidence values (0.98) in contrast with the low confidence value reported by SSD++ (0.348).

\item In terms of accuracy prediction, regarding the TIC-TAC-TOE dataset, all algorithms report an accuracy value around 0.57. However, with respect to VOTE and P4Lucat datasets, both IGSD versions and FSSD report a high accuracy value above 0.8 in VOTE dataset for IGSD algorithm and above 0.7 in P4Lucat for the mentioned algorithms. These confidence values show the high reliability of IGSD and FSSD for VOTE and P4Lucat datasets. Finally, considering the GENBASE dataset, all the algorithms are highly reliable due to reported accuracy values above 0.9.

\item Considering the average IG per rule, with respect to TIC-TAC-TOE and VOTE datasets, IGSD-M and IGSD-T report high IG values (0.072, 0.45) and (0.072, 0.33), respectively, and SSD++ reports a high IG value considering the VOTE dataset (0.41). Furthermore, regarding the P4Lucat dataset, SSD++ was able to discover patterns with an IG value of 0.034, while slightly lower values are reported by both IGSD versions (0.016, 0.011) and FSSD (0.019), respectively. Finally, considering the GENBASE dataset, FSSD, SSD++ and IGSD-T produce sets of patterns with a similar IG value (0.1, 0.15, 0.1), respectively, while IGSD-M set of patterns has a lower IG value of 0.048.

\item Taking into consideration the ORR, patterns produced by IGSD-M and IGSD-T report the highest ORR value for all the datasets. On the other hand, FSSD reports an ORR value slightly lower in TIC-TAC-TOE dataset, and a significantly lower ORR value regarding P4Lucat.
In addition, although SSD++ reports the same ORR value as the rest of the algorithms considering the GENBASE dataset, regarding the other datasets SSD++ reports a considerably lower ORR value.

\item In terms of p-value, regarding TIC-TAC-TOE and GENBASE  datasets, both IGSD versions and FSSD algorithms produce highly statistically significant patterns due to a p-value below 0.001. In turn, SSD++ obtains a statistically non-relevant set of patterns for TIC-TAC-TOE dataset, reporting a p-value above 0.05, while a relevant set of patterns is obtained by the same algorithm for GENBASE dataset, reporting a p-value below 0.05. Furthermore, concerning the VOTE dataset, both IGSD versions and SDD++ algorithms produce highly statistically significant patterns, reporting a p-value below 0.001, while FSSD reports a p-value above 0.05, thus providing a statistically non-relevant set of patterns for the same dataset. Finally, with respect to P4Lucat, both IGSD versions report a p-value below 0.01, meanwhile, FSSD reports a p-value below 0.05 and SDD++ reports a p-value above 0.05.
\end{itemize}

\begin{table}[H]
\centering
  \setlength\tabcolsep{6pt}
  \caption{{Statistical comparison using SD algorithms for Mixed datasets.}}
  \label{tab:resMixDatasets}
  \hfill%
  \resizebox{\textwidth}{!}{%
  \subfloat[BREAST-CANCER dataset]{%
    \label{tab:breast_stats}%
    \begin{tabular}{|ccccc|l}
\cline{1-5}
\multicolumn{5}{|c|}{\textbf{BREAST-CANCER}} &
   \\ \cline{1-5}
\multicolumn{1}{|c|}{} &
  \multicolumn{1}{c|}{\textbf{IGSD-M}} &
  \multicolumn{1}{c|}{\textbf{IGSD-T}} &
  \multicolumn{1}{c|}{\textbf{FSSD}} &
  \textbf{SSD++} &
   \\ \cline{1-5}
\multicolumn{1}{|c|}{\textbf{size}} &
  \multicolumn{1}{c|}{\cellcolor[HTML]{E9E898}4} &
  \multicolumn{1}{c|}{\cellcolor[HTML]{D3E193}5} &
  \multicolumn{1}{c|}{\cellcolor[HTML]{63BE7B}10} &
  \cellcolor[HTML]{FFEF9C}3 &
   \\ \cline{1-5}
\multicolumn{1}{|c|}{\textbf{length}} &
  \multicolumn{1}{c|}{\cellcolor[HTML]{99CF87}3} &
  \multicolumn{1}{c|}{\cellcolor[HTML]{B8D98D}2.6} &
  \multicolumn{1}{c|}{\cellcolor[HTML]{63BE7B}3.7} &
  \cellcolor[HTML]{FFEF9C}1.67 &
   \\ \cline{1-5}
\multicolumn{1}{|c|}{\textbf{coverage}} &
  \multicolumn{1}{c|}{\cellcolor[HTML]{FFEF9C}0.072} &
  \multicolumn{1}{c|}{\cellcolor[HTML]{FFEF9C}0.072} &
  \multicolumn{1}{c|}{\cellcolor[HTML]{63BE7B}0.11} &
  \cellcolor[HTML]{8DCB84}0.1 &
   \\ \cline{1-5}
\multicolumn{1}{|c|}{\textbf{wracc}} &
  \multicolumn{1}{c|}{\cellcolor[HTML]{63BE7B}0.03} &
  \multicolumn{1}{c|}{\cellcolor[HTML]{63BE7B}0.03} &
  \multicolumn{1}{c|}{\cellcolor[HTML]{B3D88C}0.02} &
  \cellcolor[HTML]{FFEF9C}0.0085 &
   \\ \cline{1-5}
\multicolumn{1}{|c|}{\textbf{confidence}} &
  \multicolumn{1}{c|}{\cellcolor[HTML]{63BE7B}0.91} &
  \multicolumn{1}{c|}{\cellcolor[HTML]{9ACF87}0.82} &
  \multicolumn{1}{c|}{\cellcolor[HTML]{9ACF87}0.82} &
  \cellcolor[HTML]{FFEF9C}0.65 &
   \\ \cline{1-5}
\multicolumn{1}{|c|}{\textbf{accuracy}} &
  \multicolumn{1}{c|}{\cellcolor[HTML]{B1D78C}0.56} &
  \multicolumn{1}{c|}{\cellcolor[HTML]{63BE7B}0.66} &
  \multicolumn{1}{c|}{\cellcolor[HTML]{B1D78C}0.56} &
  \cellcolor[HTML]{FFEF9C}0.46 &
   \\ \cline{1-5}
\multicolumn{1}{|c|}{\textbf{info\_gained}} &
  \multicolumn{1}{c|}{\cellcolor[HTML]{91CD85}0.053} &
  \multicolumn{1}{c|}{\cellcolor[HTML]{91CD85}0.053} &
  \multicolumn{1}{c|}{\cellcolor[HTML]{FFEF9C}0.024} &
  \cellcolor[HTML]{63BE7B}0.065 &
   \\ \cline{1-5}
\multicolumn{1}{|c|}{\textbf{odd-range}} &
  \multicolumn{1}{c|}{\cellcolor[HTML]{63BE7B}4} &
  \multicolumn{1}{c|}{\cellcolor[HTML]{83C882}3.8} &
  \multicolumn{1}{c|}{\cellcolor[HTML]{F0EB99}3.1} &
  \cellcolor[HTML]{FFEF9C}3 &
   \\ \cline{1-5}
\multicolumn{1}{|c|}{\textbf{p-value}} &
  \multicolumn{1}{c|}{\cellcolor[HTML]{69C07C}5.50E-03} &
  \multicolumn{1}{c|}{\cellcolor[HTML]{67BF7C}4.20E-03} &
  \multicolumn{1}{c|}{\cellcolor[HTML]{FFEF9C}1.20E-01} &
  \cellcolor[HTML]{63BE7B}4.78E-04 &
   \\ \cline{1-5}
\end{tabular}}
  \hfill%
  \subfloat[NURSERY dataset]{%
    \label{tab:nursery_stats}%
    \begin{tabular}{|ccccc|l}
\cline{1-5}
\multicolumn{5}{|c|}{\textbf{NURSERY}} &
   \\ \cline{1-5}
\multicolumn{1}{|c|}{} &
  \multicolumn{1}{c|}{\textbf{IGSD-M}} &
  \multicolumn{1}{c|}{\textbf{IGSD-T}} &
  \multicolumn{1}{c|}{\textbf{FSSD}} &
  \textbf{SSD++} &
   \\ \cline{1-5}
\multicolumn{1}{|c|}{\textbf{size}} &
  \multicolumn{1}{c|}{\cellcolor[HTML]{FFEF9C}13} &
  \multicolumn{1}{c|}{\cellcolor[HTML]{F2EB9A}20} &
  \multicolumn{1}{c|}{\cellcolor[HTML]{FFEF9C}13} &
  \cellcolor[HTML]{63BE7B}97 &
   \\ \cline{1-5}
\multicolumn{1}{|c|}{\textbf{length}} &
  \multicolumn{1}{c|}{\cellcolor[HTML]{D9E394}2.85} &
  \multicolumn{1}{c|}{\cellcolor[HTML]{FFEF9C}2.65} &
  \multicolumn{1}{c|}{\cellcolor[HTML]{63BE7B}3.46} &
  \cellcolor[HTML]{CADF91}2.93 &
   \\ \cline{1-5}
\multicolumn{1}{|c|}{\textbf{coverage}} &
  \multicolumn{1}{c|}{\cellcolor[HTML]{F4EC9A}0.063} &
  \multicolumn{1}{c|}{\cellcolor[HTML]{B1D78C}0.08} &
  \multicolumn{1}{c|}{\cellcolor[HTML]{63BE7B}0.1} &
  \cellcolor[HTML]{FFEF9C}0.06 &
   \\ \cline{1-5}
\multicolumn{1}{|c|}{\textbf{wracc}} &
  \multicolumn{1}{c|}{\cellcolor[HTML]{63BE7B}0.039} &
  \multicolumn{1}{c|}{\cellcolor[HTML]{84C982}0.033} &
  \multicolumn{1}{c|}{\cellcolor[HTML]{94CE86}0.03} &
  \cellcolor[HTML]{FFEF9C}0.01 &
   \\ \cline{1-5}
\multicolumn{1}{|c|}{\textbf{confidence}} &
  \multicolumn{1}{c|}{\cellcolor[HTML]{63BE7B}0.87} &
  \multicolumn{1}{c|}{\cellcolor[HTML]{C5DD90}0.72} &
  \multicolumn{1}{c|}{\cellcolor[HTML]{F9ED9B}0.64} &
  \cellcolor[HTML]{FFEF9C}0.63 &
   \\ \cline{1-5}
\multicolumn{1}{|c|}{\textbf{accuracy}} &
  \multicolumn{1}{c|}{\cellcolor[HTML]{63BE7B}0.78} &
  \multicolumn{1}{c|}{\cellcolor[HTML]{63BE7B}0.78} &
  \multicolumn{1}{c|}{\cellcolor[HTML]{E0E696}0.74} &
  \cellcolor[HTML]{FFEF9C}0.73 &
   \\ \cline{1-5}
\multicolumn{1}{|c|}{\textbf{info\_gained}} &
  \multicolumn{1}{c|}{\cellcolor[HTML]{63BE7B}0.092} &
  \multicolumn{1}{c|}{\cellcolor[HTML]{93CD86}0.07} &
  \multicolumn{1}{c|}{\cellcolor[HTML]{A9D48A}0.06} &
  \cellcolor[HTML]{FFEF9C}0.02 &
   \\ \cline{1-5}
\multicolumn{1}{|c|}{\textbf{odd-range}} &
  \multicolumn{1}{c|}{\cellcolor[HTML]{63BE7B}4} &
  \multicolumn{1}{c|}{\cellcolor[HTML]{7CC681}3.8} &
  \multicolumn{1}{c|}{\cellcolor[HTML]{CADF91}3.15} &
  \cellcolor[HTML]{FFEF9C}2.7 &
   \\ \cline{1-5}
\multicolumn{1}{|c|}{\textbf{p-value}} &
  \multicolumn{1}{c|}{\cellcolor[HTML]{63BE7B}4.48E-133} &
  \multicolumn{1}{c|}{\cellcolor[HTML]{63BE7B}3.47E-133} &
  \multicolumn{1}{c|}{\cellcolor[HTML]{63BE7B}1.01E-12} &
  \cellcolor[HTML]{FFEF9C}8.00E-03 &
   \\ \cline{1-5}
\end{tabular}}%
}
  \hspace{\fill}

  \vspace{4pt}
  
  \hfill%
  
  \resizebox{2.75in}{!}{%
  \subfloat[HEART dataset]{%
    \label{tab:heart_stats}
    
    \begin{tabular}{|ccccc|l}
\cline{1-5}
\multicolumn{5}{|c|}{\textbf{HEART}} &
   \\ \cline{1-5}
\multicolumn{1}{|c|}{} &
  \multicolumn{1}{c|}{\textbf{IGSD-M}} &
  \multicolumn{1}{c|}{\textbf{IGSD-T}} &
  \multicolumn{1}{c|}{\textbf{FSSD}} &
  \textbf{SSD++} &
   \\ \cline{1-5}
\multicolumn{1}{|c|}{\textbf{size}} &
  \multicolumn{1}{c|}{\cellcolor[HTML]{D8E394}7} &
  \multicolumn{1}{c|}{\cellcolor[HTML]{63BE7B}22} &
  \multicolumn{1}{c|}{\cellcolor[HTML]{FFEF9C}2} &
  \cellcolor[HTML]{E0E696}6 &
   \\ \cline{1-5}
\multicolumn{1}{|c|}{\textbf{length}} &
  \multicolumn{1}{c|}{\cellcolor[HTML]{FFEF9C}1.86} &
  \multicolumn{1}{c|}{\cellcolor[HTML]{FBEE9C}2.14} &
  \multicolumn{1}{c|}{\cellcolor[HTML]{63BE7B}13} &
  \cellcolor[HTML]{FFEF9C}1.83 &
   \\ \cline{1-5}
\multicolumn{1}{|c|}{\textbf{coverage}} &
  \multicolumn{1}{c|}{\cellcolor[HTML]{9DD088}0.257} &
  \multicolumn{1}{c|}{\cellcolor[HTML]{DDE595}0.22} &
  \multicolumn{1}{c|}{\cellcolor[HTML]{FFEF9C}0.2} &
  \cellcolor[HTML]{63BE7B}0.29 &
   \\ \cline{1-5}
\multicolumn{1}{|c|}{\textbf{wracc}} &
  \multicolumn{1}{c|}{\cellcolor[HTML]{63BE7B}0.089} &
  \multicolumn{1}{c|}{\cellcolor[HTML]{87CA83}0.075} &
  \multicolumn{1}{c|}{\cellcolor[HTML]{93CE86}0.07} &
  \cellcolor[HTML]{FFEF9C}0.027 &
   \\ \cline{1-5}
\multicolumn{1}{|c|}{\textbf{confidence}} &
  \multicolumn{1}{c|}{\cellcolor[HTML]{63BE7B}0.87} &
  \multicolumn{1}{c|}{\cellcolor[HTML]{6BC17D}0.86} &
  \multicolumn{1}{c|}{\cellcolor[HTML]{89CA83}0.82} &
  \cellcolor[HTML]{FFEF9C}0.66 &
   \\ \cline{1-5}
\multicolumn{1}{|c|}{\textbf{accuracy}} &
  \multicolumn{1}{c|}{\cellcolor[HTML]{63BE7B}0.67} &
  \multicolumn{1}{c|}{\cellcolor[HTML]{8BCB84}0.64} &
  \multicolumn{1}{c|}{\cellcolor[HTML]{98CF87}0.63} &
  \cellcolor[HTML]{FFEF9C}0.55 &
   \\ \cline{1-5}
\multicolumn{1}{|c|}{\textbf{info\_gained}} &
  \multicolumn{1}{c|}{\cellcolor[HTML]{B1D78C}0.137} &
  \multicolumn{1}{c|}{\cellcolor[HTML]{EAE998}0.11} &
  \multicolumn{1}{c|}{\cellcolor[HTML]{FFEF9C}0.1} &
  \cellcolor[HTML]{63BE7B}0.174 &
   \\ \cline{1-5}
\multicolumn{1}{|c|}{\textbf{odd-range}} &
  \multicolumn{1}{c|}{\cellcolor[HTML]{63BE7B}4} &
  \multicolumn{1}{c|}{\cellcolor[HTML]{63BE7B}4} &
  \multicolumn{1}{c|}{\cellcolor[HTML]{B1D78C}3.5} &
  \cellcolor[HTML]{FFEF9C}3 &
   \\ \cline{1-5}
\multicolumn{1}{|c|}{\textbf{p-value}} &
  \multicolumn{1}{c|}{\cellcolor[HTML]{63BE7B}1.37E-09} &
  \multicolumn{1}{c|}{\cellcolor[HTML]{63BE7B}5.13E-07} &
  \multicolumn{1}{c|}{\cellcolor[HTML]{FFEF9C}2.30E-04} &
  \cellcolor[HTML]{63BE7B}2.62E-08 &
   \\ \cline{1-5}
\end{tabular}}%
}
  
  \hspace{\fill}
\end{table}

Table~\ref{tab:resMixDatasets} shows the obtained metric values for the mixed datasets considered. Also, table~\ref{tab:resMixDatasets} uses a color range (from green to yellow) to indicate the best and worst metric value for each row, being thick green the best and thick yellow the worst. The summary of results in terms of descriptive and predictive measures is as follows:
\begin{itemize}
\item Regarding the BREAST-CANCER dataset, FSSD returns a higher number of discovered patterns (10) with respect to both IGSD versions and SSD++ algorithms, which return a similar amount of patterns (4, 5, 3), respectively. Furthermore, concerning the NURSERY dataset, SSD++ was able to discover a higher amount of patterns (97). In turn, a lower number of patterns is returned by both IGSD versions (13, 20) and FSSD (13), respectively. Finally, with respect to the HEART dataset, IGSD-T discover more patterns (22) than IGSD-M, FSSD and SSD++ algorithms (7, 2, 6) respectively.
\item In terms of the complexity of the rule set (i.e. length), FSSD produces sets of patterns with the largest number of variables for the 3 datasets. In addition, it can be noticed that regarding the HEART dataset, FSSD set of patterns has a much higher number of variables (13) than both IGSD versions (1.86, 2.14) and SSD++ (1.83), respectively.
\item Regarding the average coverage per rule, considering the BREAST-CANCER and NURSERY datasets, patterns produced by FSSD have slightly higher values (0.11, 0.1), respectively, than IGSD-M (0.072, 0.063), IGSD-T (0.072, 0.08) and SSD++ (0.1, 0.06). In turn, SSD++ reports for the HEART dataset the highest coverage value (0.29), followed by both IGSD versions (0.257, 0.22) and FSSD (0.2).
\item Assessing the unusualness (i.e. WRacc) of the patterns for BREAST-CANCER, NURSERY and HEART datasets, IGSD-M obtains sets of patterns with the highest unusualness values (0.03, 0.039, 0.089), respectively. Moreover, IGSD-T and FSSD report slightly lower values (0.03, 0.033, 0.075) and (0.02, 0.03, 0.07), respectively, while SSD++ reports much lower values (0.0085, 0.01, 0.027).
\item When evaluating the rule confidence for BREAST-CANCER, NURSERY and HEART datasets, IGSD-M produces sets of patterns with the highest confidence values (0.91, 0.87, 0.87). Furthermore, IGSD-T and FSSD report slightly lower values (0.82, 0.72, 0.86) and (0.82, 0.64, 0.82), respectively, while SSD++ reports much lower values compared to the others (0.65, 0.63, 0.66).
\item In terms of accuracy prediction, regarding the NURSERY dataset, all the algorithms might be considered reliable models due to the reported accuracy values (0.78, 0.78, 0.74, and 0.73) respectively for each algorithm. Regarding the BREAST-CANCER and HEART datasets, the accuracy values decrease to values around 0.6, although SSD++ reports low accuracy values (0.46, 0.55) respectively. 
\item Considering the average IG per rule, SSD++ was able to discover patterns in BREAST-CANCER and HEART datasets with the highest IG value (0.065, 0.174), respectively, while slightly lower IG values are reported for patterns discovered by IGSD-M (0.53, 0.53) and IGSD-T (0.137, 0.11), respectively. In turn, FSSD is the worst performant reporting the lowest IG values for all three datasets. 
\item Regarding ORR values, patterns produced by IGSD-M report the highest ORR value (4) for all the datasets, while patterns produced by IGSD-T have ORR values slightly lower for BREAST-CANCER and NURSERY datasets. On the other hand, FSSD and SDD++ algorithms report lower ORR values regarding all the datasets (3.1, 3.15, 3.5) and (3, 2.7, 3), respectively.  
\item In terms of p-value, both IGSD versions and SDD++ algorithms produce statistically significant patterns for BREAST-CANCER dataset, reporting a p-value below 0.01. In turn, FSSD algorithm provides non-significant patterns reporting a p-value above 0.05. Furthermore, concerning the NURSERY and HEART datasets, all the algorithms report a p-value below 0.001, with the exception of SSD++ algorithm in the NURSERY dataset, reporting a p-value above 0.001; which still indicates statistical significance.
\end{itemize}

Analyzing all metric results presented above, it can be seen that patterns generated by IGSD and SSD++ have similar complexity values, while FSSD produces patterns with more complexity and amount of information. However, considering the nominal TIC-TAC-TOE and P4Lucat datasets, IGSD was able to produce patterns with more complexity than FSSD and SSD++. Additionally, the sets returned by IGSD usually contain more patterns than FSSD and SSD++. Thus, it can be concluded that IGSD produces larger sets of patterns with less amount of information or variables for each returned pattern. \\

Considering the coverage measure, there is variability among different dataset types, all three algorithms can discover patterns with considerable representation in datasets, but both IGSD methods were successful for all datasets while FSSD and SSD++ failed for one numeric dataset. Looking at coverage and ORR measures together, it can be noticed that as a general rule, IGSD patterns are more reliable due to reported values, being the patterns produced by SSD++ the least reliable. In addition, 
the p-value of the patterns produced by IGSD are always below 0.05, being in several datasets below 0.001, thus these patterns can be considered as highly statistically significant. On the other hand, FSSD and SSD++ can not guarantee statistically significant patterns due to p-value scores above 0.05 reported for IRIS, TIC-TAC-TOE, VOTE, P4LUCAT, and BREAST-CANCER datasets. Therefore, in summary, it can be concluded that IGSD is able to discover a considerable number of statistically significant patterns with also a high dependence on targets, but offering less amount of information than FSSD.
On the other hand, it is noticeable that for numeric datasets such as MAGIC and ECHO, FSSD and SSD++, executed with default parameter values, failed to discover patterns, while IGSD was able to finish and uncover a set of patterns. This limitation could be due to the way FSSD and SSD++ handle the generation of ranges using numeric data. \\

Regarding pattern search exploration, IGSD managed to provide patterns for all datasets but SSD++ and FSSD failed to find any pattern for ECHO and MAGIC datasets, respectively. However, GENBASE dataset, with the highest number of columns (1186), made IGSD reach the one-hour time limit while searching for patterns. This is due to the fact that both FSSD and SSD++ follow a greedy search strategy, using list of subgroups, and IGSD uses sets of subgroups which allow to do not discard too early potentially good patterns. This exploration strategy makes IGSD to require longer computational times but enables a larger exploration of the search space, allowing to obtain potentially more relevant patterns in this way.


\subsection{Experts validation}
\label{sec:Validation}
This section presents the results of the validation performed by domain experts for the P4Lucat dataset and discusses the interrelation between the evaluation and the pattern measures results obtained in Section~\ref{sec:Results1}. A total of 92 patterns, containing the output of IGSD, FSSD, and SSD++, were given to the group of expert raters. \\

We first assess the quality of the clinical validation by calculating AC1 and ICC indices for the validated patterns discovered by all algorithms. Table \ref{tab:indicesraters} contains indices values showing a moderate agreement regarding AC1. On the other hand, ICC shows moderate reliability. 

\begin{table}[H]
\centering
\resizebox{0.5\textwidth}{!}{%
\begin{tabular}{|c|c|c|}
\hline
\textbf{Index} & \textbf{AC1} & \multicolumn{1}{c|}{\textbf{ICC}} \\ \hline
\textbf{Value} & 0.48 & 0.60 \\ \hline
\textbf{CI} & (0.35, 0.58) & (0.46,   0.71) \\ \hline
\textbf{p-value} & 3,09E-12 & 3,89E-08 \\ \hline
\end{tabular}%
}
\caption{Indices of raters for pattern validation, CI: Confidence interval.}
\label{tab:indicesraters}
\end{table}

Table~\ref{tab:validationacceptance} shows the evaluator's acceptance rate values for the patterns provided by each algorithm using the P4Lucat dataset. It can be seen that IGSD-M achieved the highest average acceptance rate. Overall, Table~\ref{tab:validationacceptance} shows that each of the two IGSD methods provides significantly higher average acceptance rates and also achieves better results on an evaluator basis.
\begin{table}[H]
\centering
\resizebox{\textwidth}{!}{%
\begin{tabular}{|c|c|c|c|c|c|c|c|c|}
\hline
\textbf{Algorithm/Users} &
  \textbf{Rater1} &
  \textbf{Rater2} &
  \textbf{Rater3} &
  \textbf{Rater4} &
  \textbf{Rater5} &
  \textbf{Rater6} &
  \textbf{Rater7} &
  \textbf{Average} \\ \hline
\textbf{IGSD-T} &
  \cellcolor[HTML]{63BE7B}64\% &
  \cellcolor[HTML]{F8696B}10\% &
  \cellcolor[HTML]{72C37C}56\% &
  \cellcolor[HTML]{63BE7B}40\% &
  \cellcolor[HTML]{63BE7B}27\% &
  \cellcolor[HTML]{F8696B}10\% &
  \cellcolor[HTML]{FAEA84}33\% &
  \cellcolor[HTML]{AED480}34\% \\ \hline
\textbf{IGSD-M} &
  \cellcolor[HTML]{95CD7E}58\% &
  \cellcolor[HTML]{63BE7B}26\% &
  \cellcolor[HTML]{63BE7B}58\% &
  \cellcolor[HTML]{8ECB7E}37\% &
  \cellcolor[HTML]{E3E383}16\% &
  \cellcolor[HTML]{63BE7B}37\% &
  \cellcolor[HTML]{63BE7B}47\% &
  \cellcolor[HTML]{63BE7B}40\% \\ \hline
\textbf{SSD++} &
  \cellcolor[HTML]{F8696B}0\% &
  \cellcolor[HTML]{FDD680}18\% &
  \cellcolor[HTML]{F8696B}0\% &
  \cellcolor[HTML]{F8696B}0\% &
  \cellcolor[HTML]{F8696B}0\% &
  \cellcolor[HTML]{FDD680}18\% &
  \cellcolor[HTML]{F8696B}18\% &
  \cellcolor[HTML]{F8696B}8\% \\ \hline
\textbf{FSSD} &
  \cellcolor[HTML]{FCC57C}32\% &
  \cellcolor[HTML]{DCE182}21\% &
  \cellcolor[HTML]{FBA276}16\% &
  \cellcolor[HTML]{FDC77D}21\% &
  \cellcolor[HTML]{FDD27F}11\% &
  \cellcolor[HTML]{F2E884}21\% &
  \cellcolor[HTML]{FEE683}32\% &
  \cellcolor[HTML]{FCC37C}22\% \\ \hline
\end{tabular}%
}
\caption{Acceptance rate of patterns in different algorithms}
\label{tab:validationacceptance}
\end{table}

When comparing validation and performance results provided in Section~\ref{sec:Results1} for P4Lucat dataset, the method with higher acceptance rates (i.e. IGSD) also provided significantly higher values of standard metrics from SD literature like confidence and higher values of non-standard performance metrics used in this work, like ORR and p-value. 


Looking at the rest of datasets, it can be seen that IGSD, in general, provides higher confidence and ORR values than FSSD and SSD++, while providing a p\_value below 0.05. Thus, although unfortunately validation was not possible to be performed in this work for the rest of datasets, based on the validation of the P4Lucat dataset, we consider that the use of non-standard SD performance metrics like: IG, ORR and p-value can complement standard SD metrics and allow to better evaluate discovered patterns.


%% file: tex/conclusions.tex

\section{Conclusions}

In this work, we have proposed Information Gained Subgroup Discovery (IGSD), a new SD algorithm for pattern discovery that combines Information Gain and Odds Ratio as a multi-criteria for pattern selection. Additionally, two versions of IGSD are proposed to evaluate the dynamic adjustment of the search optimization thresholds during subgroup space exploration. Also, main and general limitations of state-of-the-art SD algorithms are discussed, identifying the following ones: need for fine-tuning of key parameters for each dataset, usage of a single pattern search criteria set by hand, usage of non-overlapping data structures for subgroup space exploration, and impossibility to search for patterns by fixing some relevant dataset variables. The proposed IGSD algorithm tries to tackle all these limitations and thus is evaluated using up to eleven datasets with different characteristics to uncover patterns. For comparison purposes, the same datasets are also used with two state-of-the-art SD algorithms: FSSD and SSD++.

Results obtained showed that FSSD provides more complex patterns and SSD++ provides less complex patterns than IGSD. In turn, IGSD usually finds more larger patterns sets than FSSD and SSD++. Thus, it can be concluded that IGSD produces larger sets of patterns with less amount of information or variables for each returned pattern. On the other hand, FSSD and SSD++ confidence average values are 83\% and 63\%, respectively, significantly lower than IGSD confidence average values of around 90\%. This lower reliability of FSSD and SSD++ is also reflected in ORR average values providing 3.58 and 3, respectively, stating a medium-high dependence between patterns and targets. In turn, IGSD provided an ORR average value of around 4, stating a high dependence between patterns and targets. The fact that IGSD obtained better results than FSSD and SSD++ without manual setting of any search parameter also validates the proposed method.

In the performance evaluation of patterns obtained by compared algorithms for all datasets, we propose to complement standard SD measures and include some metrics: Information Gain, ORR and p\-value, not considered typically in SD literature. Also, results obtained for P4Lucat dataset have been validated by a group of experts. Thus, patterns acceptance rates show that results provided by IGSD, are more in agreement with the experts than results obtained using FSSD and SSD++ algorithms. For the P4Lucat dataset, better-accepted patterns also have higher ORR and confidence values while being statistically significant with a p\-value below 0.05. Hence, we consider that the inclusion of the proposed non-standard SD metrics allows to better evaluate discovered patterns. 

Finally, as mentioned above, the proposed IGSD algorithm uses sets of subgroups and follows a non-greedy pattern search strategy. This makes IGSD perform a wider exploration of the search space, allowing to obtain potentially more relevant patterns but at the cost of significantly longer computational times. As a future work, we plan to explore a similar strategy we adopted in a previous work \cite{CBMS23} for selecting statistically significant variables. Then, the set of variables in datasets with a large number of columns can be reduced, and search patterns based on this reduced set of significant variables.